\documentclass[hyphens]{article}
\usepackage[final]{neurips_2025} 

\usepackage[utf8]{inputenc}
\usepackage[T1]{fontenc}
\usepackage{hyperref}
\hypersetup{
 colorlinks=true,
 linkcolor=red,
 citecolor=cyan,
 filecolor=magenta,
 urlcolor=magenta,
}
\usepackage{url}
\usepackage{xurl}
\usepackage{enumitem}
\usepackage{booktabs}
\usepackage{amsfonts}
\usepackage{nicefrac}
\usepackage{microtype}
\usepackage[table,dvipsnames]{xcolor}
\usepackage{amsmath}
\usepackage{cleveref}
\usepackage{xspace}
\usepackage{textcomp}
\usepackage{stfloats}
\usepackage{verbatim}
\usepackage{wrapfig}
\usepackage{graphicx}

\usepackage{fancyhdr}
\renewcommand{\headwidth}{\textwidth}

\renewcommand{\headrulewidth}{0.5pt}
\renewcommand{\headrule}{%
    \vspace{2pt}%
    \hbox to\headwidth{%
        \color{black}%
        \leaders\hrule height \headrulewidth\hfill
    }%
}

\usepackage{float}
\usepackage{amssymb}
\usepackage[numbers,sort&compress]{natbib}
\usepackage{tikz}
\usepackage{algorithm}
\usepackage{algpseudocode}
\usepackage{makecell}
\usepackage{multicol,multirow}
\usepackage{threeparttable}
\usepackage{tablefootnote}
\usepackage{pgfplots}
\pgfplotsset{compat=1.18}
\usepackage[labelfont=bf]{caption}
\usepackage{subcaption}
\definecolor{yarnblue}{RGB}{225,238,250}
\usepackage{array}

\AtBeginEnvironment{thebibliography}{\raggedright\sloppy}
\hfuzz=\maxdimen
\vfuzz=\maxdimen

\title{TuringLLM: Efficiently Scaling Foundation Models Toward Physical AI}

\author{
\colorbox{white}{\includegraphics[height=0.8em]{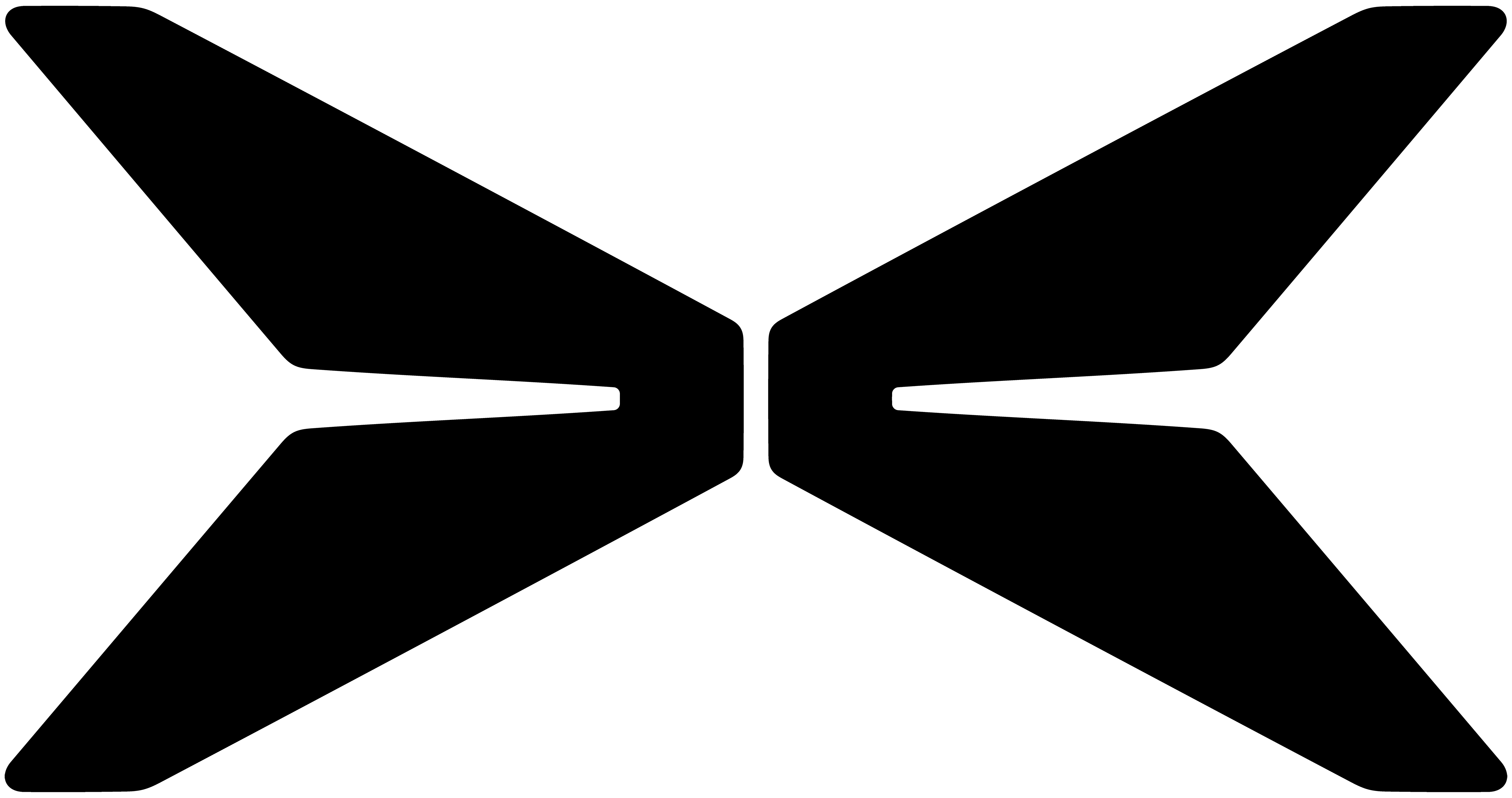}}
Foundation Model Team, Xpeng Inc.
}

\begin{document}
\setlength{\headheight}{22.4pt}
\addtolength{\topmargin}{-10.4pt}
\maketitle
\begin{abstract}
We present \textbf{Turing-20B-A2B}, a 20B-parameter Mixture-of-Experts
language model that activates approximately 2B parameters per token,
designed for long-context and latency-sensitive physical AI
applications. The model adopts Quantile Routing in a dynamic top-$k$
configuration, enabling token-adaptive expert allocation while
maintaining balanced expert utilization and a controlled average
compute budget. During deployment, we further apply
capacity-constrained routing to prompt prefill for more regular and
efficient expert execution, while retaining dropless routing during
pretraining. Turing-20B-A2B also employs a
hybrid attention architecture that combines Lightning Attention with a
small number of full-attention layers for efficient long-context
modeling. The model is pretrained with a progressive three-stage
curriculum and extended to a native context length of 128K through
continued pretraining, with further inference-time extension to 512K
using YaRN. Despite its compact active-parameter budget,
Turing-20B-A2B achieves, at the base-model stage, overall general
capability exceeding Qwen3-8B Base and approaching Qwen3.5-9B Base,
while maintaining strong long-context performance and favorable
prefill-latency scaling. These results demonstrate an effective balance
among model capability, long-context scalability, and practical
inference efficiency.
\end{abstract}

\begin{figure}[H]
    \centering
    \includegraphics[width=1.0\textwidth]
    {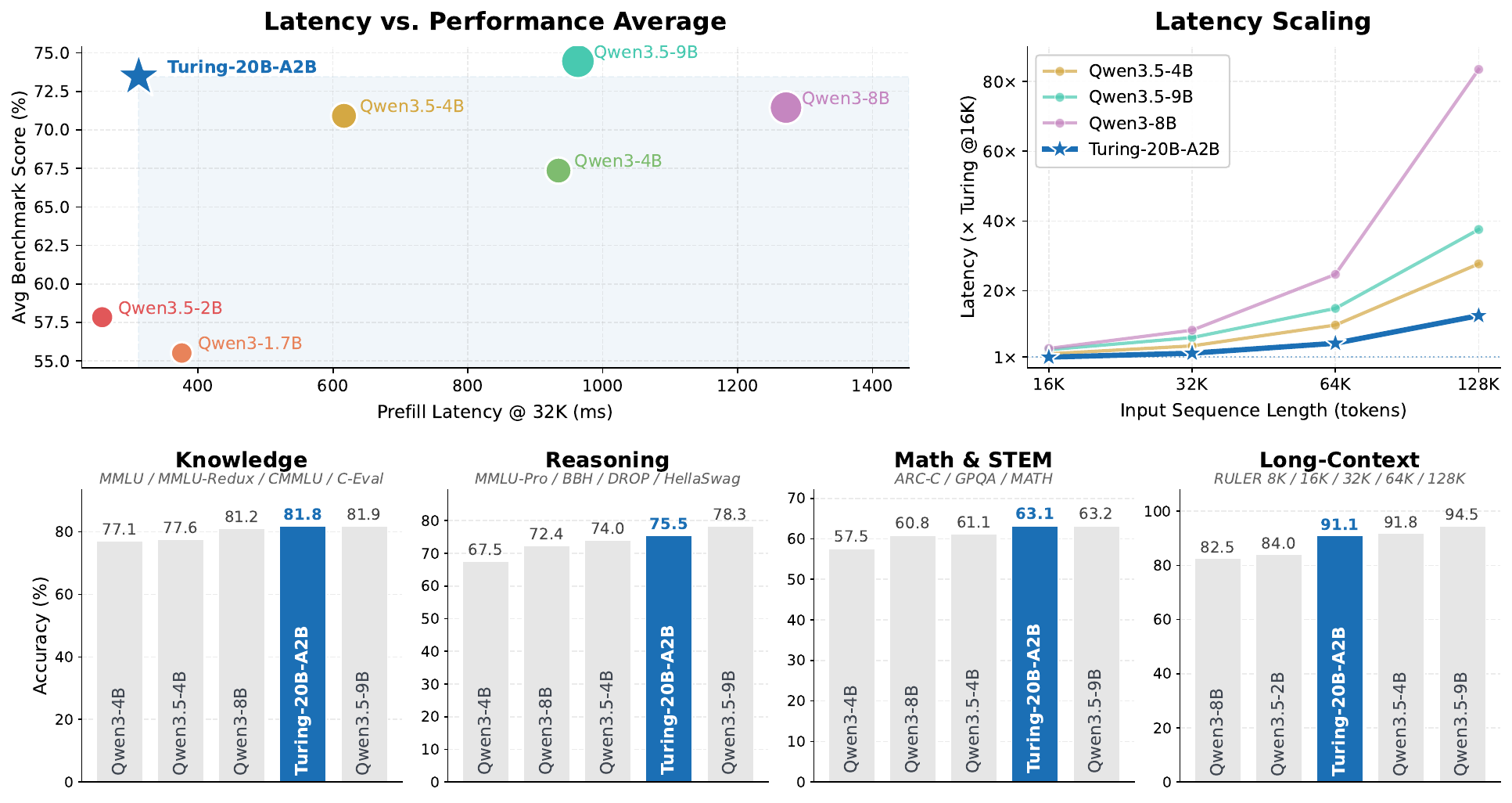}
    \caption{
        \textbf{Overall capability and inference-efficiency comparison of
        Turing-20B-A2B with representative open-source base models.}
        The overall benchmark score averages all evaluated Knowledge,
        Reasoning, and Math \& STEM benchmarks. Prefill latency is
        measured on NVIDIA H800 GPUs.
    }
    \label{fig:overall_comparison}
\end{figure}
\section{Introduction}
\label{sec:introduction}

In physical AI systems such as autonomous driving and embodied
intelligence, large language models (LLMs) are increasingly used as
foundation models to support perception, reasoning, decision-making,
and the prediction of action-conditioned environment
dynamics~\cite{driess2023palme,brohan2023rt2,kim2024openvla,
shao2023lmdrive,hu2023gaia1}. These systems require the underlying
model to balance several competing objectives. To make reliable
decisions in complex physical environments, the model must retain
broad world knowledge and strong general reasoning capabilities.
Meanwhile, foundation models deployed in autonomous-driving and
embodied systems increasingly need to process extended histories of
visual observations, system states, and actions to capture temporal
dependencies and support coherent behavior over long
interactions~\cite{shao2023lmdrive,xue2024longvila,
liu2023ringattention}. Such long-context inputs substantially increase
the computational and memory costs of model
inference~\cite{dao2023flashattention2,chen2023longlora}, while
closed-loop interaction requires decisions to be produced under strict
latency constraints. Therefore, a practical foundation model for
physical AI must simultaneously provide strong general capabilities,
efficient long-context modeling, and low-latency inference.

Scaling model capacity has been a major driver of capability
improvements in modern LLMs~\cite{kaplan2020scaling,
hoffmann2022training}. However, larger dense models generally require
more computation for each token, increasing inference cost and response
latency in latency-sensitive physical AI systems. Mixture-of-Experts
(MoE) architectures alleviate part of this tension by increasing total
model capacity while keeping the number of parameters activated per
token relatively small~\cite{fedus2022switch}. MoE architectures have
consequently become a common design choice in recent frontier
open-weight models, including DeepSeek-V4, Qwen3.6, and Kimi
K3~\cite{deepseekv4,qwen36,kimik3}. Nevertheless, computational
sparsity does not automatically translate into proportional inference
speedups: conventional top-$k$ routing can produce imbalanced expert
loads and irregular execution, making efficient deployment dependent
on additional system-level optimization~\cite{wang2024lossfree,
he2025capacity}. At the same time, the long observation and action
histories common in physical AI applications make standard full
attention increasingly expensive in both computation and memory.
Recent frontier models therefore adopt efficient or hybrid
attention architectures to improve long-context efficiency while
preserving expressive token interactions~\cite{deepseekv4,qwen35,
kimik3}. Despite these advances, many frontier models still operate
with relatively large active-parameter budgets or require substantial
deployment optimization to realize their architectural efficiency in
practice. Achieving strong general capabilities, favorable long-context
latency scaling, and efficient deployment within a compact
active-parameter budget therefore remains challenging.

To address these challenges, we develop
\textbf{Turing-20B-A2B}, a 20B-parameter Mixture-of-Experts language
model that activates approximately 2B parameters per token on average.
The model is designed around a deployment-oriented scaling recipe for
physical AI foundation models. It combines dynamic top-$k$ Quantile
Routing~\cite{su2025adaptive_moe} for token-adaptive expert allocation,
a hybrid attention backbone dominated by Lightning
Attention~\cite{qin2024lightning2} for efficient long-context
processing, progressive continued pretraining for context extension,
and capacity-constrained expert execution during prompt prefill for
more regular and predictable deployment behavior. Quantile Routing
allows different tokens to activate different numbers of experts while
maintaining balanced expert utilization and a controlled average
compute budget through online quantile tracking. The hybrid attention
design substantially reduces the contribution of quadratic-complexity
attention as context length grows while retaining periodic global token
interactions through full-attention layers. During pretraining, MoE
execution remains dropless, whereas deployment-time expert-capacity
control bounds per-expert workloads to improve the regularity and
efficiency of prompt prefill. Progressive long-context continued
pretraining further extends the native context window from 4K to 128K,
followed by inference-time extension to 512K using YaRN. Together,
these components provide a practical path to scaling foundation-model
capability and context length without proportionally increasing
per-token computation or sacrificing deployment efficiency.

The effectiveness of this design is reflected in the model's
empirical performance. As shown in
Figure~\ref{fig:overall_comparison}, at the base-model stage,
Turing-20B-A2B outperforms Qwen3-8B Base in the overall benchmark
average and approaches the substantially larger Qwen3.5-9B Base,
despite activating only approximately 2B parameters per token on
average. Across capability categories, Turing-20B-A2B exhibits
particularly strong performance on knowledge and math \& STEM
benchmarks while remaining competitive in reasoning and long-context
evaluations. Meanwhile, the model maintains favorable prefill-latency
scaling over long contexts, with the efficiency advantage over
representative Qwen3 and Qwen3.5 baselines becoming increasingly
pronounced from 32K to 128K on NVIDIA H800 GPUs with FP16 precision.
Together, these results demonstrate that the proposed
deployment-oriented scaling recipe achieves a favorable balance among
model capability, long-context scalability, and inference efficiency
under a compact active-parameter budget, making Turing-20B-A2B well
suited to latency-sensitive physical AI workloads that require both
extended context processing and efficient execution.
\begin{figure*}[t]
    \centering

    \begin{minipage}[t]{0.68\textwidth}
        \vspace{0pt}
        \centering
        \includegraphics[
            width=\linewidth,
            trim=15 0 15 20,
            clip
        ]{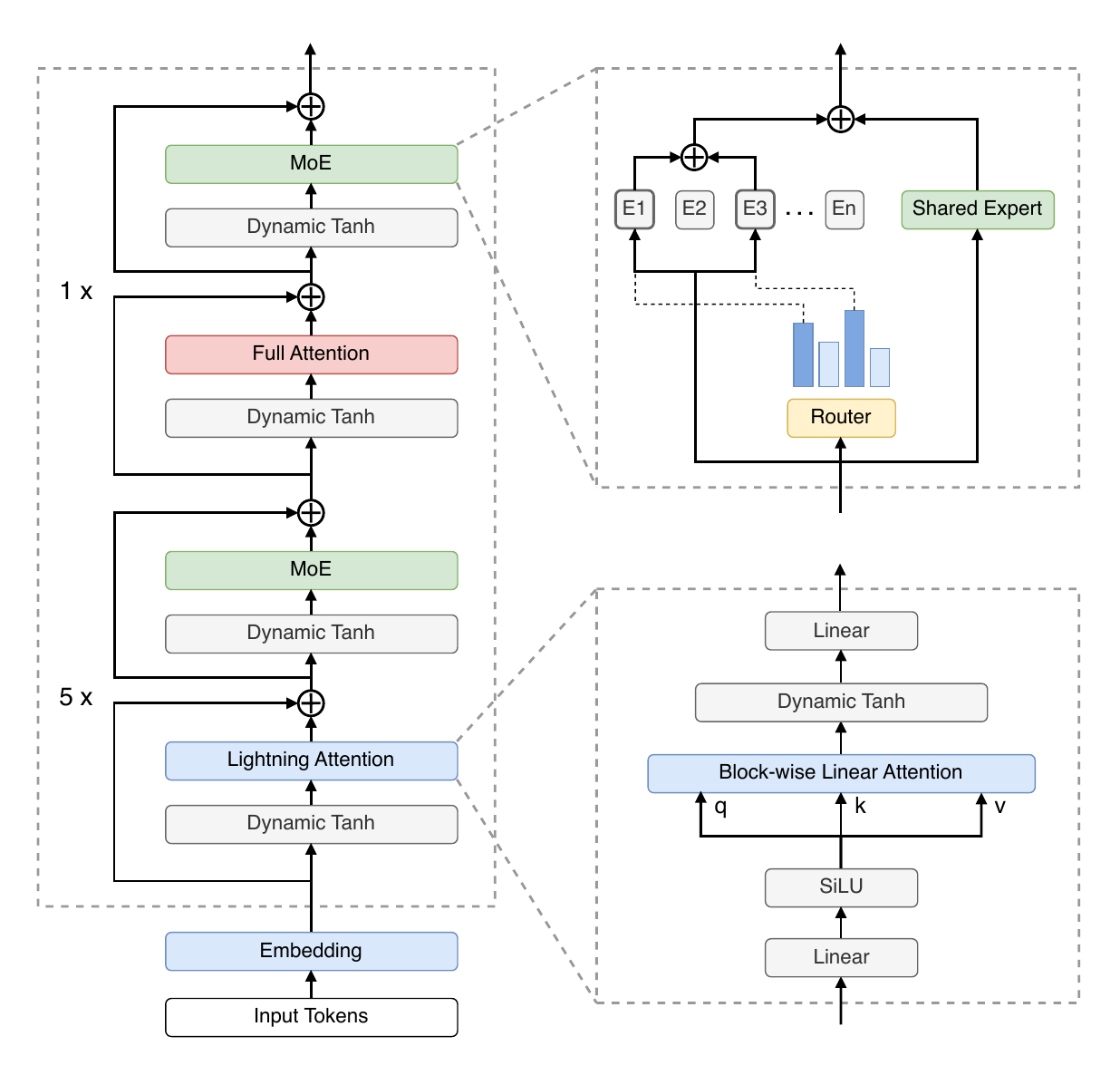}
    \end{minipage}
    \hfill
    \begin{minipage}[t]{0.31\textwidth}
        \vspace{1.6em}
        \centering
        \small
        \renewcommand{\arraystretch}{1.08}
        \setlength{\tabcolsep}{3pt}

        \begin{tabular}{@{\hspace{3pt}}lr@{\hspace{3pt}}}
            \toprule
            \multicolumn{2}{c}{\textbf{Turing-20B-A2B}} \\
            \midrule

            \rowcolor{gray!10}
            \multicolumn{2}{l}{\textbf{Model}} \\
            Total parameters       & 20B \\
            Activated parameters   & $\sim$2B \\
            Layers                 & 24 \\
            Hidden size            & 2,048 \\
            Native context         & 128K \\
            Extended context       & 512K \\[2pt]

            \rowcolor{gray!10}
            \multicolumn{2}{l}{\textbf{MoE}} \\
            Routed experts         & 256 \\
            Active experts         & $\sim$8 \\
            Shared experts         & 1 \\
            Routed expert dim      & 512 \\
            Shared expert dim      & 2,048 \\[2pt]

            \rowcolor{gray!10}
            \multicolumn{2}{l}{\textbf{Attention}} \\
            Attention heads        & 16 \\
            Head dimension         & 128 \\
            Lightning Attn. layers & 20 \\
            Full Attn. layers      & 4 \\

            \bottomrule
        \end{tabular}
    \end{minipage}

    \caption{
        \textbf{Overview of the Turing-20B-A2B architecture and model
        configuration.}
        The decoder stack alternates groups of five Lightning Attention
        layers with one full-attention layer. The table summarizes
        the main architectural configurations.
    }

    \label{fig:overall_architecture}
\end{figure*}

\section{Model Architecture}
\label{sec:methodology}

Turing-20B-A2B is a decoder-only Mixture-of-Experts language model
designed to balance model capacity, long-context efficiency, and
inference cost under a compact activated-parameter budget. The model
contains approximately 20B parameters in total while activating
approximately 2B parameters per token. As illustrated in
Figure~\ref{fig:overall_architecture}, Turing-20B-A2B combines a
hybrid-attention backbone with sparse MoE feed-forward modules and
token-adaptive expert routing. Most decoder layers employ Lightning
Attention for efficient long-context computation, while periodic
full-attention layers preserve global token interactions. The
feed-forward modules are predominantly implemented as sparse MoE
layers with routed experts and a shared expert.

\subsection{Overall Architecture}
\label{sec:overall_architecture}

Turing-20B-A2B consists of 24 decoder layers with a hidden dimension
of 2,048. A central consideration in the architectural design is
deployment simplicity. Because the target inference stack includes
internal edge devices with a relatively constrained operator set, we
favor modules that can be implemented with simple and efficient
primitives and avoid introducing unnecessary architectural
complexity. This consideration motivates the use of the basic
Lightning Attention formulation for linear-attention layers, as well
as Dynamic Tanh (DyT)~\cite{zhu2025transformers} in place of
conventional RMSNorm. Together, these choices keep the core decoder
structure lightweight and deployment-friendly while preserving the
modeling capacity required for large-scale pretraining and
long-context extension.

The decoder stack follows a repeating hybrid-attention pattern. Every
group of six layers contains five Lightning Attention layers followed
by one full-attention layer, resulting in 20 Lightning Attention
layers and four full-attention layers in total. Each attention module
is preceded by DyT, and its output is combined with the residual stream
before entering the feed-forward module. This design substantially
reduces the fraction of layers that incur quadratic attention cost
while retaining periodic global token interactions through full
attention.

The first decoder layer uses a dense feed-forward network, while the
remaining layers use sparse MoE feed-forward modules. Each MoE module
contains 256 routed experts and one shared expert. The routed experts
use an intermediate dimension of 512, whereas the shared expert uses
an intermediate dimension of 2,048. Approximately eight routed
experts are activated per token on average, allowing the model to
increase total parameter capacity while maintaining a compact
per-token computation budget. The detailed MoE formulation and
expert-routing strategy are described in
Sections~\ref{sec:moe} and~\ref{sec:routing_strategy}, respectively.

The final model supports a native context length of 128K tokens and is
further extended to 512K at inference time using YaRN. The following
subsections describe the hybrid attention mechanism, sparse MoE
architecture, and routing strategy in detail.

\subsection{Hybrid Attention}
\label{sec:hybrid_attention}

Turing-20B-A2B adopts a hybrid attention architecture in which
Lightning Attention~\cite{qin2024lightning2} is used in the majority
of decoder layers and standard causal full attention is inserted
periodically. As described in Section~\ref{sec:overall_architecture},
the model follows a $5{:}1$ pattern, with five Lightning Attention
layers followed by one full-attention layer. This design reduces the
computational growth associated with long-context processing while
retaining periodic global token interactions.

Our Lightning Attention implementation largely follows the formulation
introduced in Lightning Attention-2~\cite{qin2024lightning2}. The
attention computation is performed in a block-wise linear-attention
form, combining intra-block attention with recurrent inter-block state
propagation. We additionally enable the exponential decay mechanism,
with a different decay rate assigned to each attention head. For head
$h$, the interaction between positions $i$ and $j$ is modulated by
\begin{equation}
    D_h(i,j)
    =
    \exp\left(-s_h(i-j)\right),
    \qquad i \geq j,
\end{equation}
where $s_h$ is a head-specific decay slope. The slopes are constructed
using an ALiBi-style head-wise schedule~\cite{press2022train}, so that
different attention heads operate with different effective temporal
ranges. Equivalently, by defining $\lambda_h=\exp(-s_h)$, the decay can
be written as $D_h(i,j)=\lambda_h^{\,i-j}$. This block-wise formulation
avoids explicitly materializing the full quadratic attention matrix and
therefore provides substantially more favorable scaling with sequence
length.

Periodic full-attention layers complement Lightning Attention by
providing unrestricted token-to-token interactions across the complete
context. In Turing-20B-A2B, four of the 24 decoder layers use causal
full attention, while the remaining 20 layers use Lightning Attention.
This hybrid design provides a practical balance between long-context
efficiency and global information exchange.

\subsection{Mixture-of-Experts}
\label{sec:moe}

To increase model capacity without proportionally increasing the
per-token computation cost, Turing-20B-A2B replaces the dense
feed-forward networks in most decoder layers with sparse
Mixture-of-Experts (MoE) modules. The first decoder layer retains a
dense feed-forward network, while the remaining layers employ sparse
MoE blocks. In practice, expert loads in the earliest MoE layer are
often more difficult to balance, since token representations are still
at a relatively low level of abstraction. We therefore use a dense
feed-forward layer at the bottom of the network and introduce sparse
expert routing only in subsequent layers.

Each MoE block contains 256 routed experts together with one shared
expert. For an input token representation $\mathbf{x}_t$, the router
produces expert scores and selects a sparse set of routed experts
$\mathcal{S}_t$. The routed branch is computed as
\begin{equation}
    \mathbf{y}^{\mathrm{routed}}_t
    =
    \sum_{e \in \mathcal{S}_t}
    w_{t,e}\,
    \operatorname{FFN}_e(\mathbf{x}_t),
\end{equation}
where $w_{t,e}$ denotes the routing weight assigned to expert $e$.
In parallel, every token is processed by a shared expert,
$\operatorname{FFN}_{\mathrm{shared}}$, and the final MoE output is
given by
\begin{equation}
    \mathbf{y}_t
    =
    \mathbf{y}^{\mathrm{routed}}_t
    +
    \operatorname{FFN}_{\mathrm{shared}}(\mathbf{x}_t).
\end{equation}

The routed experts use an intermediate dimension of 512, whereas the
shared expert uses a substantially larger intermediate dimension of
2,048. Besides providing a dense computation path shared by all
tokens, the larger shared expert also serves as a stronger fallback
path for the capacity-constrained routing used during prompt prefill.
When some routed-expert assignments are removed because an expert
exceeds its capacity, the shared expert remains available to every
token and helps preserve a stable minimum amount of feed-forward
computation. We describe the capacity-constrained routing mechanism in
Section~\ref{sec:routing_strategy}.


\begin{figure*}[t]
    \centering
    \includegraphics[width=1.0\textwidth]
    {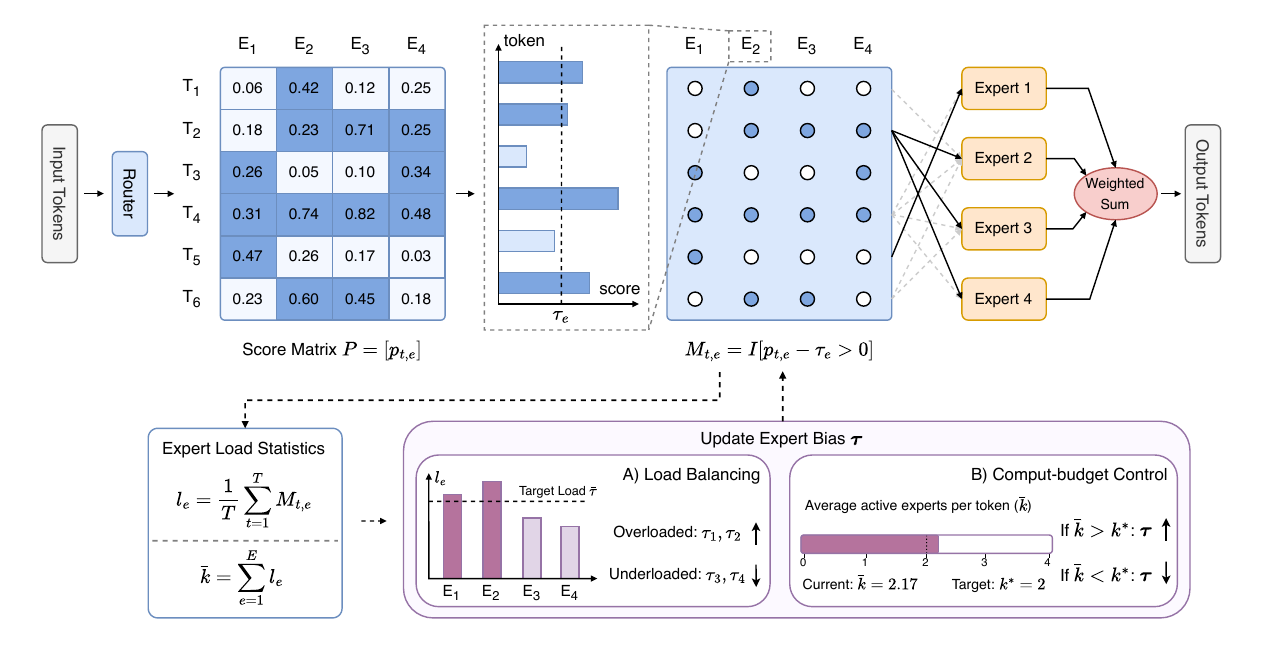}
    \caption{
        \textbf{Overview of the routing strategy.}
        Expert-specific thresholds determine token--expert assignments and are
        updated from expert-load statistics during training to track the target
        $\left(1-k^{*}/E\right)$ score quantiles. This unified quantile-tracking
        process naturally balances expert utilization, controls the average
        routing budget, and allows token-adaptive expert activation.
    }
    \label{fig:quantile_routing}
\end{figure*}

\subsection{Routing Strategy}
\label{sec:routing_strategy}

Turing-20B-A2B adopts Quantile Routing~\cite{su2025adaptive_moe}
in a dynamic top-$k$ configuration. Unlike conventional top-$k$
routing, which assigns every token to a fixed number of experts,
this configuration determines expert activation by comparing router
scores against expert-specific thresholds. As a result, different
tokens may activate different numbers of routed experts according to
their score distributions, while the average routing budget remains
controlled around a target value.

As illustrated in Figure~\ref{fig:quantile_routing}, given a token
representation $\mathbf{x}_t$, the router produces a score
$p_{t,e}$ for each routed expert $e$. Each expert is associated with
an expert-specific threshold $\tau_e$, and the corresponding
token--expert assignment is defined as
\begin{equation}
    M_{t,e}
    =
    \mathbb{I}\left[p_{t,e} > \tau_e\right],
    \label{eq:quantile_mask}
\end{equation}
where $M_{t,e}\in\{0,1\}$ indicates whether token $t$ is dispatched
to expert $e$. The number of routed experts activated by token $t$ is
therefore
\begin{equation}
    k_t
    =
    \sum_{e=1}^{E} M_{t,e},
\end{equation}
which is allowed to vary across tokens.

The key idea of Quantile Routing is to adapt each expert-specific
threshold toward a target quantile of its router-score distribution.
Given a target average routing budget of $k^{*}$ experts per token and
$E$ routed experts, the desired activation probability of each expert
is approximately $k^{*}/E$. Accordingly, the threshold $\tau_e$ is
driven toward the
$\left(1-k^{*}/E\right)$ quantile of the score distribution
$\{p_{t,e}\}_{t=1}^{T}$ for expert $e$, such that
\begin{equation}
    \Pr(p_{t,e} > \tau_e)
    \approx
    \frac{k^{*}}{E}.
\end{equation}
Intuitively, this places approximately the highest-scoring
$k^{*}/E$ fraction of tokens above the routing threshold for each
expert.

We track the target quantile through expert-load statistics and corresponding
bias updates. For expert $e$, the normalized load is
\begin{equation}
    \ell_e
    =
    \frac{1}{T}
    \sum_{t=1}^{T} M_{t,e}.
    \label{eq:expert_load}
\end{equation}
As shown in Figure~\ref{fig:quantile_routing}, the bias-update
procedure incorporates both the relative load of individual experts
and the current average number of activated experts. Together, they form
the adjustment that moves the expert-specific thresholds toward
their target quantiles.

Once the thresholds approach the target, expert balancing
and computation-budget control arise naturally from the same routing
mechanism. Since each expert accepts approximately a $k^{*}/E$
fraction of tokens, expert loads tend toward a balanced distribution.
At the same time, the expected number of routed experts activated per
token satisfies
\begin{equation}
    \mathbb{E}[k_t]
    =
    \sum_{e=1}^{E}
    \Pr(p_{t,e} > \tau_e)
    \approx
    k^{*}.
\end{equation}
Thus, Quantile Routing simultaneously provides balanced expert
utilization and control of the average expert budget through a unified
quantile-tracking process, while allowing the model to adaptively
allocate different amounts of expert computation to tokens of varying
difficulty. 
In Turing-20B-A2B, we set the target average routing budget
to approximately eight routed experts per token.

During pretraining, we keep MoE routing dropless and do not impose an
explicit expert-capacity constraint~\cite{fedus2022switch}. This avoids
introducing capacity-induced competition among tokens, particularly for
packed training sequences containing multiple independent documents.
Moreover, global score-based capacity selection may cause the expert
assignment of an earlier token to depend on router scores from later tokens
in the same packed sequence, potentially introducing a non-causal dependency
into the MoE execution path. We therefore retain dropless routing throughout
these training stages. For prompt prefill during reinforcement learning and
deployment, however, we additionally impose an expert-capacity constraint to
bound the maximum workload assigned to individual experts.

For an input containing $T$ prefill tokens, the capacity of each
routed expert is defined as
\begin{equation}
    C_e
    =
    \left\lceil
        \gamma \frac{T k^{*}}{E}
    \right\rceil,
    \qquad \gamma = 1.25,
    \label{eq:expert_capacity}
\end{equation}
where $\gamma$ denotes the capacity factor. When the number of
assignments routed to an expert exceeds its capacity, the assignments
are ranked according to their router scores and only the highest-scoring
$C_e$ assignments are retained~\cite{zhou2022expert}. 
This score-based overflow policy
preferentially preserves assignments with stronger routing preferences
while explicitly bounding the maximum workload of each expert.

During reinforcement learning, each device processes a single
QA sample, so capacity competition remains within one prompt and does
not couple independent samples. Moreover, because the complete prompt serves
as observed conditioning context, score-based selection does not access future
response tokens and therefore preserves the causal structure of
autoregressive generation. During deployment, bounding the per-expert
workload also makes the expert-execution path more regular and improves
long-context prefill efficiency. We evaluate the effect of this design
on model capability and MoE-module prefill latency in
Section~\ref{sec:capacity_ablation}.

\section{Training Recipe}
\label{sec:training_recipe}

\begin{figure*}[t]
    \centering
    \includegraphics[width=0.92\textwidth]{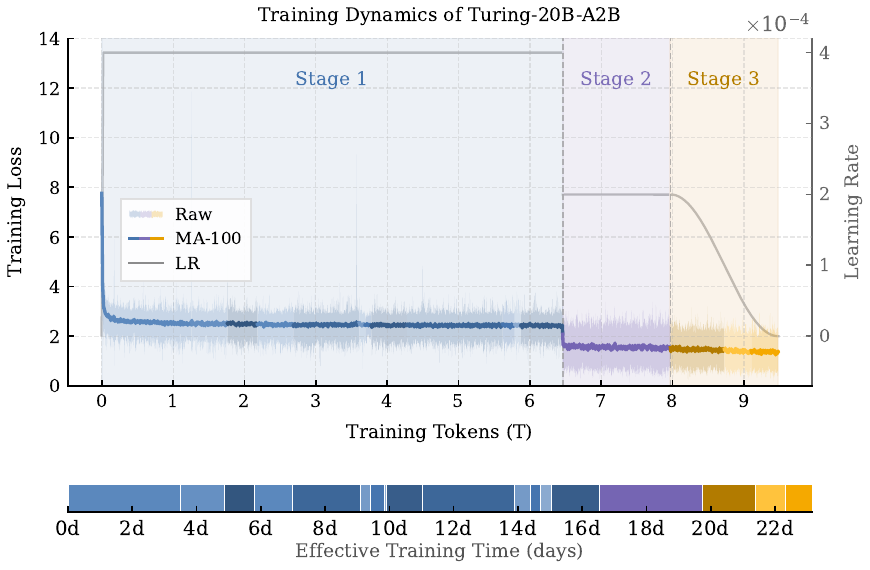}
    \caption{
        \textbf{Training dynamics of Turing-20B-A2B over the three-stage main
        pretraining curriculum.} The upper panel shows the raw training loss,
        its 100-step moving average (MA-100), and the learning-rate schedule
        over processed tokens, with dashed lines indicating stage boundaries.
        The lower timeline shows the effective training time of individual
        resumed runs.
    }
    \label{fig:training_dynamics}
\end{figure*}

We train Turing-20B-A2B using a progressive curriculum designed to
first establish broad language and knowledge capabilities, then
strengthen capability-dense domains, and finally consolidate the model
through quality-focused annealing. The main pretraining process is
organized into three stages---\emph{knowledge foundation},
\emph{capability enhancement}, and \emph{quality annealing}---with
stage-dependent data mixtures and learning-rate schedules while keeping
the model architecture and training objective unchanged. After the main $4$K-context pretraining curriculum, we further extend the native
context window through continued long-context training from $4$K to $32$K and
subsequently to $128$K. Overall, the main pretraining process was completed in
approximately 22 days using 512 NVIDIA H800 GPUs.
Figure~\ref{fig:training_dynamics} summarizes the corresponding training
trajectory, including the training loss, learning-rate schedule, stage
transitions, and effective training time.

\subsection{Pretraining Data}
\label{sec:pretraining_data}

Our pretraining corpus is primarily bilingual in English and Chinese
and covers a broad range of domains, including general web text,
mathematics, source code, scientific content, and other
knowledge-intensive data. The data composition is progressively
adjusted across the three-stage curriculum. Stage~1 emphasizes
large-scale and diverse natural corpora to establish broad linguistic
and knowledge coverage. Stage~2 increases the contribution of
higher-quality filtered and capability-dense data, while Stage~3
further shifts toward internally constructed data for capability
consolidation. We do not disclose the exact mixture ratios of
individual data sources.

For general web data, we use large-scale public corpora such as
DCLM~\cite{li2024datacomplm} and
FineWeb-Edu~\cite{penedo2024fineweb} during the foundation stage,
followed by higher-quality filtered English and Chinese web data in
the later stages. The filtering and data-selection pipelines for web
and mathematical corpora follow the quality-oriented tiered data
management methodology explored in
UltraData~\cite{wang2026ultradata}. In particular, we apply
deduplication, cleaning, and quality-aware selection to reduce noisy,
redundant, and low-information content while preserving broad topical
coverage.

Mathematical data includes
FineMath~\cite{allal2025smollm2},
InfiMM-WebMath~\cite{han2025infimmwebmath}, and
OpenWebMath~\cite{paster2023openwebmath} in Stage~1, with
MegaMath~\cite{zhou2025megamath} and additional filtered mathematical
data introduced in Stage~2. Stage~3 further incorporates in-house
mathematical data. For code, we primarily use
The Stack~v2~\cite{lozhkov2024starcoder2} and
Stack-Edu~\cite{allal2025smollm2} during Stages~1 and~2, while
Stage~3 uses in-house code data. Knowledge-oriented data includes
peS2o~\cite{soldaini2023pes2o}, Wikipedia, and
Cosmopedia~v2~\cite{allal2025smollm2} in Stage~1, followed by
in-house knowledge data in the later stages. A relatively small
amount of book and academic-paper data is additionally introduced
during Stages~2 and~3 to improve coverage of long-form, structured,
and specialized knowledge.

Our in-house corpora consist of both internally processed natural data
and synthetic data. Synthetic data primarily targets mathematics,
code, and knowledge, with the goal of increasing the diversity and
density of high-value concepts, structured reasoning patterns, and
specialized knowledge that may be underrepresented in naturally
occurring corpora. Such data plays a larger role in the later stages,
where training increasingly emphasizes data quality and capability
density over raw corpus scale.

Table~\ref{tab:pretraining_data} summarizes the representative data
sources used across the three stages. Overall, the data curriculum
progressively transitions from broad natural coverage toward
higher-quality filtered and in-house data, providing a broad
foundation in the early stage while allowing the later stages to
focus more strongly on capability-dense training signals.

\begin{table*}[t]
    \centering
    \small
    \renewcommand{\arraystretch}{1.12}
    \setlength{\tabcolsep}{4pt}
    \caption{
        Representative data sources used across the three-stage
        pretraining curriculum. Public datasets are listed by name,
        while filtered and in-house entries denote internally processed
        or constructed corpora. Exact mixture ratios are not disclosed.
    }
    \label{tab:pretraining_data}

    \begin{tabular*}{0.94\textwidth}{
        @{\extracolsep{\fill}}
        >{\raggedright\arraybackslash}m{0.11\textwidth}
        >{\centering\arraybackslash}m{0.22\textwidth}
        >{\centering\arraybackslash}m{0.25\textwidth}
        >{\centering\arraybackslash}m{0.26\textwidth}
        @{}
    }
        \toprule
        \textbf{Category}
            & \textbf{S1: Knowledge Foundation}
            & \textbf{S2: Capability Enhancement}
            & \textbf{S3: Quality Annealing} \\
        \midrule

        \textbf{Web}
            & DCLM, FineWeb-Edu
            & filtered web data
            & filtered web data \\

        \cmidrule{1-4}

        \textbf{Math}
            & FineMath, InfiWebMath, OpenWebMath
            & MegaMath, filtered math data
            & in-house math data \\

        \cmidrule{1-4}

        \textbf{Code}
            & The Stack~v2, Stack-Edu
            & The Stack~v2, Stack-Edu
            & in-house code data \\

        \cmidrule{1-4}

        \textbf{Knowledge}
            & peS2o, Wikipedia, Cosmopedia~v2
            & in-house knowledge data, books/papers
            & in-house knowledge data, books/papers \\

        \bottomrule
    \end{tabular*}
\end{table*}

\subsection{Optimization and Training Curriculum}
\label{sec:training_curriculum}

The main pretraining process follows a three-stage curriculum while
sharing a common optimization and distributed-training setup. We use a
sequence length of $4{,}096$ and a global batch size of $2{,}048$
sequences, corresponding to approximately $8.4$M tokens per
optimization step. Training is performed with BF16 mixed precision
using AdamW with $\beta_1=0.9$, $\beta_2=0.95$,
$\epsilon=10^{-8}$, and a weight decay of $0.1$. Weight decay is not
applied to embedding parameters, and gradients are clipped to a maximum
norm of $1.0$. In addition to the standard next-token prediction
objective, we adopt Multi-Token Prediction (MTP) following
DeepSeek-V3~\cite{deepseekv3}, using a single additional prediction
head ($N_{\mathrm{MTP}}=1$) with a loss weight of
$\lambda_{\mathrm{MTP}}=0.1$. For distributed MoE training, we use expert parallelism
with degree $8$, while tensor and pipeline parallelism are both set to
$1$. A distributed optimizer is used, with gradient reduction performed
in FP32.

For all training data, we adopt length-aware bin packing following
Ding et al.~\cite{ding2024fewer} to improve sequence utilization and
reduce unnecessary truncation and padding. The treatment of attention
across packed documents evolves with the curriculum: Stage~1 uses
standard causal attention over the packed sequence, whereas Stages~2
and~3 additionally apply a document-level attention mask to prevent
tokens from attending across document boundaries.

As illustrated in Figure~\ref{fig:training_dynamics}, the main
pretraining process is organized into three progressive stages:
\emph{knowledge foundation}, \emph{capability enhancement}, and
\emph{quality annealing}. The model architecture and training objective
remain unchanged throughout the curriculum, while the data distribution,
learning-rate schedule, and packed-sequence attention treatment are
progressively adjusted. Stage~1 runs for $770$K optimization steps
($\sim6.5$T tokens), followed by $180$K steps in Stage~2 and another
$180$K steps in Stage~3 (approximately $1.5$T tokens each).

\paragraph{Stage 1: knowledge foundation.}
Training begins with a $2{,}000$-step warm-up to a peak learning rate
of $4\times10^{-4}$, which is then kept constant throughout Stage~1.
The data mixture emphasizes large-scale and diverse natural data to
establish broad linguistic coverage, world knowledge, and general
modeling capability.

\paragraph{Stage 2: capability enhancement.}
At the transition to Stage~2, the learning rate is reduced to
$2\times10^{-4}$ and held constant. The data distribution shifts toward
higher-quality and more capability-dense content, with increased
emphasis on mathematics, code, and specialized knowledge while retaining
filtered web data for general knowledge. Starting from this stage,
a document-level attention mask is applied to packed sequences to
preserve document independence.

\paragraph{Stage 3: quality annealing.}
Stage~3 continues from the same learning rate of $2\times10^{-4}$ and
applies cosine decay to $10^{-6}$. The training mixture further shifts
toward higher-quality in-house data, with greater emphasis on
capability-dense mathematics, code, and knowledge data. The
document-level attention mask introduced in Stage~2 is retained
throughout this stage.

Overall, the curriculum progressively shifts the training emphasis from
broad coverage to capability density and finally to data quality, while
the learning-rate schedule transitions from high-rate foundation
training to low-rate quality-focused annealing. 

\subsection{Long-Context Training}
\label{sec:long_context}

After completing the three-stage main pretraining curriculum at a
context length of $4$K, we progressively extend the native context
window of Turing-20B-A2B through two continued-pretraining stages,
first from $4$K to $32$K and subsequently from $32$K to $128$K.
Rather than directly training at the target context length, this
progressive schedule gradually exposes the model to increasingly
longer sequences while adjusting the positional encoding and
context-parallel configuration accordingly. Both stages are trained
on $256$ NVIDIA H800 GPUs with a global token batch of approximately
$8.4$M tokens per optimization step. Each stage runs for $5{,}000$
steps with a peak learning rate of $8\times10^{-5}$, a $150$-step
warm-up, and a WSD schedule with cosine decay to a minimum learning
rate of $10^{-6}$.

\paragraph{Long-context data mixture.}
Both continued-pretraining stages use the same dedicated long-context
data mixture. We retain $45\%$ of the Stage~3 data distribution,
preserving its original relative proportions to mitigate degradation
of capabilities acquired during main pretraining. Synthetic
long-context data accounts for another $20\%$ and includes
needle-in-a-haystack tasks, frequency-based aggregation tasks,
long-context question answering, and other synthetic formats targeting
long-range information access and reasoning. The remaining $35\%$ consists of naturally long data, including
long and extra-long books, textbooks, and web documents derived from
ProLong~\cite{gao2025prolong} and FineWeb~\cite{penedo2024fineweb},
long-context instruction data from LongAlign~\cite{bai2024longalign},
and software-engineering data from
Scale-SWE-Distilled~\cite{zhao2026immersion}. This mixture
balances capability retention, explicit long-context training signals,
and exposure to naturally occurring long sequences.

\begin{table*}[t]
    \centering
    \small
    \renewcommand{\arraystretch}{1.12}
    \setlength{\tabcolsep}{5pt}
    \caption{
        Long-context training and inference-extension configuration.
        Turing-20B-A2B is progressively continued-pretrained from $4$K
        to $32$K and then to $128$K. The resulting $128$K checkpoint
        is further extended to $512$K with YaRN without additional
        parameter updates.
    }
    \label{tab:long_context}

    \begin{tabular*}{0.88\textwidth}{
        @{\extracolsep{\fill}}
        lccc
        @{}
    }
        \toprule
        \textbf{Configuration}
            & \textbf{LC-1}
            & \textbf{LC-2}
            & \textbf{Inference Extension} \\
        \midrule

        Context window
            & $4$K $\rightarrow$ $32$K
            & $32$K $\rightarrow$ $128$K
            & $128$K $\rightarrow$ $512$K \\

        RoPE $\theta$
            & $10^{6}$
            & $5\times10^{6}$
            & $5\times10^{6}$ \\

        Training steps
            & $5{,}000$
            & $5{,}000$
            & $0$ \\

        Context parallelism
            & $2$
            & $8$
            & -- \\

        YaRN factor
            & --
            & --
            & $4$ \\

        \bottomrule
    \end{tabular*}
\end{table*}

\paragraph{LC-1: $4$K$\rightarrow32$K.}
Starting from the final checkpoint of the main pretraining curriculum,
LC-1 increases the maximum sequence length to $32{,}768$ tokens and
raises the RoPE base to $\theta=10^{6}$. Context parallelism of degree
$2$ is used to support the longer sequence length. The resulting $32$K
checkpoint is then used to initialize the subsequent long-context
training stage.

\paragraph{LC-2: $32$K$\rightarrow128$K.}
LC-2 further increases the maximum sequence length to $131{,}072$
tokens and raises the RoPE base from $10^{6}$ to $5\times10^{6}$.
To accommodate the fourfold increase in context length, the
context-parallel degree is increased from $2$ to $8$, while the
remaining optimization configuration and global token batch are kept
unchanged. The naturally long portion of the training mixture further
incorporates long and extra-long documents to increase exposure to
sequences approaching the target context length. The resulting
checkpoint is the final model directly trained with a context window
of $128$K.

\paragraph{Inference extension: $128$K$\rightarrow512$K.}
For context lengths beyond the window observed during continued
pretraining, we apply YaRN~\cite{peng2023yarn} directly to the final
$128$K checkpoint at inference time without performing any additional
parameter updates. We retain the RoPE base
$\theta=5\times10^{6}$ and apply a scaling factor of $4$, extending
the positional range from $131{,}072$ to $524{,}288$ tokens. Thus,
$128$K is the maximum context length directly observed during
training, while the reported $256$K and $512$K results evaluate
training-free positional extrapolation.

\begin{table*}[!t]
    \centering
    \caption{
        Comparison of Turing-20B-A2B with representative Qwen3 base models
        on standard pretraining benchmarks.
        The best and second-best results are highlighted in
        \textbf{bold} and \underline{underlined}, respectively.
    }
    \label{tab:overall_performance_qwen3}

    \small
    \renewcommand{\arraystretch}{1.08}

    \begin{tabular*}{0.92\textwidth}{
        @{\extracolsep{\fill}}
        lccccc
        @{}
    }
        \toprule
        \makecell[l]{\textbf{Benchmark}}
        & \textbf{\# Shots}
        & \makecell{\textbf{Qwen3-1.7B}\\\textbf{Base}}
        & \makecell{\textbf{Qwen3-4B}\\\textbf{Base}}
        & \makecell{\textbf{Qwen3-8B}\\\textbf{Base}}
        & \makecell{\textbf{Turing-20B-A2B}} \\
        \midrule

        \multicolumn{6}{c}{\textit{Knowledge Tasks}} \\
        \midrule

        MMLU {\scriptsize (Acc.)}
        & 5-shot
        & 65.04
        & 75.44
        & \underline{78.92}
        & \textbf{79.83} \\

        MMLU-Redux {\scriptsize (Acc.)}
        & 5-shot
        & 66.83
        & 77.39
        & \textbf{81.39}
        & \underline{81.37} \\

        CMMLU {\scriptsize (Acc.)}
        & 5-shot
        & 66.46
        & 77.01
        & \underline{81.28}
        & \textbf{84.17} \\

        C-Eval {\scriptsize (Acc.)}
        & 5-shot
        & 66.90
        & 78.46
        & \textbf{83.06}
        & \underline{81.76} \\

        \midrule
        \multicolumn{6}{c}{\textit{Reasoning Tasks}} \\
        \midrule

        MMLU-Pro {\scriptsize (Acc.)}
        & 5-shot
        & 36.91
        & 50.23
        & \textbf{55.35}
        & \underline{55.17} \\

        BBH {\scriptsize (EM)}
        & 3-shot
        & 54.38
        & 70.34
        & \textbf{73.67}
        & \underline{73.61} \\

        DROP {\scriptsize (EM)}
        & 0-shot
        & 56.46
        & 76.22
        & \underline{79.03}
        & \textbf{82.61} \\

        WinoGrande {\scriptsize (Acc.)}
        & 5-shot
        & 53.83
        & 67.01
        & \textbf{70.24}
        & \underline{68.75} \\

        HellaSwag {\scriptsize (Acc.)}
        & 0-shot
        & 57.48
        & 73.16
        & \underline{81.39}
        & \textbf{90.44} \\

        \midrule
        \multicolumn{6}{c}{\textit{Math \& STEM Tasks}} \\
        \midrule

        ARC-C {\scriptsize (Acc.)}
        & 0-shot
        & 80.34
        & \underline{90.17}
        & \textbf{91.53}
        & \underline{90.17} \\

        GPQA {\scriptsize (Acc.)}
        & 5-shot
        & 24.78
        & 30.81
        & \underline{34.82}
        & \textbf{37.05} \\

        GSM8K {\scriptsize (EM)}
        & 4-shot
        & 76.95
        & \underline{89.16}
        & \textbf{90.07}
        & 74.53 \\

        MATH {\scriptsize (EM)}
        & 4-shot
        & 41.68
        & 51.56
        & \underline{56.12}
        & \textbf{62.20} \\

        \bottomrule
    \end{tabular*}
\end{table*}

\begin{table*}[!t]
    \centering
    \caption{
        Comparison of Turing-20B-A2B with representative Qwen3.5 base models
        on standard pretraining benchmarks.
        The best and second-best results are highlighted in
        \textbf{bold} and \underline{underlined}, respectively.
    }
    \label{tab:overall_performance_qwen35}

    \small
    \renewcommand{\arraystretch}{1.08}

    \begin{tabular*}{0.94\textwidth}{
        @{\extracolsep{\fill}}
        lccccc
        @{}
    }
        \toprule
        \makecell[l]{\textbf{Benchmark}}
        & \textbf{\# Shots}
        & \makecell{\textbf{Qwen3.5-2B}\\\textbf{Base}}
        & \makecell{\textbf{Qwen3.5-4B}\\\textbf{Base}}
        & \makecell{\textbf{Qwen3.5-9B}\\\textbf{Base}}
        & \makecell{\textbf{Turing-20B-A2B}} \\
        \midrule

        \multicolumn{6}{c}{\textit{Knowledge Tasks}} \\
        \midrule

        MMLU {\scriptsize (Acc.)}
        & 5-shot
        & 65.51
        & 77.53
        & \textbf{81.02}
        & \underline{79.83} \\

        MMLU-Redux {\scriptsize (Acc.)}
        & 5-shot
        & 67.24
        & 80.11
        & \textbf{82.81}
        & \underline{81.37} \\

        CMMLU {\scriptsize (Acc.)}
        & 5-shot
        & 64.46
        & 76.24
        & \underline{81.13}
        & \textbf{84.17} \\

        C-Eval {\scriptsize (Acc.)}
        & 5-shot
        & 65.05
        & 76.66
        & \textbf{82.52}
        & \underline{81.76} \\

        \midrule
        \multicolumn{6}{c}{\textit{Reasoning Tasks}} \\
        \midrule

        MMLU-Pro {\scriptsize (Acc.)}
        & 5-shot
        & 36.33
        & 52.32
        & \textbf{57.93}
        & \underline{55.17} \\

        BBH {\scriptsize (EM)}
        & 3-shot
        & 63.67
        & \underline{79.44}
        & \textbf{82.56}
        & 73.61 \\

        DROP {\scriptsize (EM)}
        & 0-shot
        & 61.66
        & 79.59
        & \textbf{83.33}
        & \underline{82.61} \\

        WinoGrande {\scriptsize (Acc.)}
        & 5-shot
        & 57.77
        & 58.25
        & \textbf{75.30}
        & \underline{68.75} \\

        HellaSwag {\scriptsize (Acc.)}
        & 0-shot
        & 64.09
        & 84.51
        & \underline{89.57}
        & \textbf{90.44} \\

        \midrule
        \multicolumn{6}{c}{\textit{Math \& STEM Tasks}} \\
        \midrule

        ARC-C {\scriptsize (Acc.)}
        & 0-shot
        & 85.76
        & \underline{93.56}
        & \textbf{94.58}
        & 90.17 \\

        GPQA {\scriptsize (Acc.)}
        & 5-shot
        & 33.04
        & \underline{37.37}
        & \textbf{41.52}
        & 37.05 \\

        GSM8K {\scriptsize (EM)}
        & 4-shot
        & 69.60
        & \underline{86.05}
        & \textbf{89.46}
        & 74.53 \\

        MATH {\scriptsize (EM)}
        & 4-shot
        & 35.76
        & 52.48
        & \underline{53.38}
        & \textbf{62.20} \\

        \bottomrule
    \end{tabular*}
\end{table*}

\section{Evaluation}\label{sec:experiments}

\subsection{Overall Performance}
\label{sec:overall_performance}
We evaluate the general capabilities of Turing-20B-A2B on thirteen widely
used benchmarks for pretrained base models, covering knowledge,
reasoning, mathematics, and STEM. For knowledge evaluation,
we use MMLU~\cite{hendrycks2021mmlu},
MMLU-Redux~\cite{gema2024mmluredux},
CMMLU~\cite{li2023cmmlu}, and C-Eval~\cite{huang2023ceval}.
For reasoning, we evaluate on
MMLU-Pro~\cite{wang2024mmlupro},
BBH~\cite{suzgun2023bbh},
DROP~\cite{dua2019drop},
WinoGrande~\cite{sakaguchi2019winogrande}, and
HellaSwag~\cite{zellers2019hellaswag}.
Math and STEM capabilities are evaluated using
ARC-Challenge (ARC-C)~\cite{clark2018arc},
GPQA~\cite{rein2023gpqa},
GSM8K~\cite{cobbe2021gsm8k}, and
MATH~\cite{hendrycks2021math}.

All models are evaluated using OpenCompass~\cite{cao2026opencompass}
under a unified generation-based evaluation protocol.
Instead of scoring candidate answers using token likelihood,
each model directly generates a textual response, from which the final
answer is extracted using a benchmark-specific post-processing
procedure. We use identical dataset versions and evaluation splits,
few-shot demonstrations and their ordering, prompt templates,
chain-of-thought settings, generation configurations, answer
extraction procedures, and scoring implementations for all compared
models. The number of in-context demonstrations and the reported
metric for each benchmark are summarized in
Tables~\ref{tab:overall_performance_qwen3}
and~\ref{tab:overall_performance_qwen35}.
Complete benchmark-specific configurations are provided in
Appendix~\ref{app:evaluation_details}.

As shown in Table~\ref{tab:overall_performance_qwen3},
Turing-20B-A2B achieves competitive overall performance compared with
Qwen3 base models despite activating only approximately 2B parameters
per token. Across the evaluated categories, the model exhibits a
balanced capability profile, with particularly strong performance in
knowledge-intensive, reasoning, and mathematical tasks. Overall,
Turing-20B-A2B reaches a performance level comparable to Qwen3-8B
while using a substantially smaller activated parameter budget per
token, demonstrating favorable parameter efficiency.

Table~\ref{tab:overall_performance_qwen35} further compares
Turing-20B-A2B with Qwen3.5 base models across multiple dense-model
scales, including 2B, 4B, and 9B variants. Although Qwen3.5-9B
achieves higher scores on a number of benchmarks, Turing-20B-A2B
remains broadly competitive across the evaluated capability categories
and preserves clear strengths on several tasks. It is worth noting
that Qwen3.5 base models may activate an explicit thinking process
during evaluation, resulting in substantially longer generations on
reasoning-intensive tasks. The comparison with Qwen3.5-2B and
Qwen3.5-4B further shows that Turing-20B-A2B maintains a strong and
well-balanced general capability profile across model scales, while
activating only approximately 2B parameters per token.

\subsection{Long-Context Ability}
\label{sec:long_context_evaluation}

We evaluate long-context capability using
RULER~\cite{hsieh2024ruler}, which assesses a range of abilities
including information retrieval, multi-hop tracing, aggregation, and
question answering over synthetically controlled context lengths.
We separately compare Turing-20B-A2B with Qwen3 and Qwen3.5 base
models to better characterize its long-context behavior across
different model generations.

Turing-20B-A2B is progressively trained to a context length of $128$K,
following the $4$K$\rightarrow32$K$\rightarrow128$K curriculum
described in Section~\ref{sec:long_context}. For evaluations beyond
the trained context window, we apply a factor-$4$ YaRN configuration
to the same $128$K checkpoint without additional continued
pretraining. In the following tables, cells marked in light blue denote
results obtained using YaRN-based context extension.

\begin{table*}[t]
    \centering
    \caption{
        Comparison of Turing-20B-A2B with representative Qwen3 base models
        on RULER across context lengths up to $128$K.
        Cells marked in {\setlength{\fboxsep}{0pt}\colorbox{yarnblue}{\hspace{1pt}light blue\hspace{1pt}}} denote
        results obtained using YaRN-based context extension.
        The best and second-best results are highlighted in
        \textbf{bold} and \underline{underlined}, respectively.
    }
    \label{tab:ruler_qwen3}

    \small
    \renewcommand{\arraystretch}{1.08}

    \begin{tabular*}{0.82\textwidth}{
        @{\hspace{\tabcolsep}\extracolsep{\fill}}
        lccccc
        @{\hspace{\tabcolsep}}
    }
        \toprule
        \multirow{2}{*}{Model}
        & \multicolumn{5}{c}{RULER} \\
        \cmidrule(lr){2-6}
        & 8K
        & 16K
        & 32K
        & 64K
        & 128K \\
        \midrule

        Qwen3-1.7B Base
        & 87.01
        & 83.67
        & 76.99
        & -
        & - \\

        Qwen3-4B Base
        & 92.40
        & 91.48
        & 85.38
        & \cellcolor{yarnblue}69.96
        & \cellcolor{yarnblue}55.77 \\

        Qwen3-8B Base
        & \textbf{94.14}
        & \underline{92.63}
        & \underline{89.26}
        & \cellcolor{yarnblue}\underline{74.83}
        & \cellcolor{yarnblue}\underline{61.56} \\

        \midrule

        \textbf{Turing-20B-A2B}
        & \underline{93.80}
        & \textbf{93.45}
        & \textbf{93.16}
        & \textbf{90.00}
        & \textbf{84.87} \\

        \bottomrule
    \end{tabular*}
\end{table*}

As shown in Table~\ref{tab:ruler_qwen3}, Turing-20B-A2B exhibits
substantially stronger scaling with context length than the Qwen3
base models. At shorter contexts, its performance is comparable to
Qwen3-8B, while the gap becomes increasingly pronounced as the
sequence length grows. Turing-20B-A2B maintains a RULER score above
90 through $64$K and retains strong performance at the trained
$128$K context limit, indicating substantially slower degradation
over long contexts.

Notably, all reported Qwen3 results beyond $32$K are obtained using
YaRN-based context extension, whereas the $64$K and $128$K results
of Turing-20B-A2B remain within its progressively trained context
range. The comparison therefore suggests that progressive
long-context training provides robust scaling toward the model's
native training limit.

\begin{table*}[t]
    \centering
    \caption{
        Comparison of Turing-20B-A2B with representative Qwen3.5 base models
        on RULER across different context lengths.
        Cells marked in {\setlength{\fboxsep}{0pt}\colorbox{yarnblue}{\hspace{1pt}light blue\hspace{1pt}}} denote
        results obtained using YaRN-based context extension.
        The best and second-best results are highlighted in
        \textbf{bold} and \underline{underlined}, respectively.
    }
    \label{tab:ruler_qwen35}

    \small
    \renewcommand{\arraystretch}{1.08}

    \begin{tabular*}{0.86\textwidth}{
        @{\hspace{\tabcolsep}\extracolsep{\fill}}
        lccccccc
        @{\hspace{\tabcolsep}}
    }
        \toprule
        \multirow{2}{*}{Model}
        & \multicolumn{7}{c}{RULER} \\
        \cmidrule(lr){2-8}
        & 8K
        & 16K
        & 32K
        & 64K
        & 128K
        & 256K
        & 512K \\
        \midrule

        Qwen3.5-2B Base
        & 92.24
        & 89.75
        & 84.02
        & 78.30
        & 75.75
        & 65.10
        & -- \\

        Qwen3.5-4B Base
        & \underline{96.72}
        & \underline{95.91}
        & \underline{93.94}
        & 89.87
        & 82.63
        & 72.56
        & -- \\

        Qwen3.5-9B Base
        & \textbf{97.72}
        & \textbf{96.97}
        & \textbf{96.05}
        & \textbf{93.01}
        & \textbf{88.89}
        & \textbf{81.81}
        & -- \\

        \midrule

        \textbf{Turing-20B-A2B}
        & 93.80
        & 93.45
        & 93.16
        & \underline{90.00}
        & \underline{84.87}
        & \cellcolor{yarnblue}\underline{81.31}
        & \cellcolor{yarnblue}\textbf{77.38} \\

        \bottomrule
    \end{tabular*}
\end{table*}

Table~\ref{tab:ruler_qwen35} provides a more challenging comparison
with the newer Qwen3.5 base models across different model scales.
At shorter context lengths, Turing-20B-A2B performs between
Qwen3.5-2B and the larger Qwen3.5-4B and Qwen3.5-9B models. As the sequence
length increases, however, Turing-20B-A2B exhibits substantially
slower performance degradation. It consistently outperforms
Qwen3.5-2B and becomes increasingly competitive with the larger
Qwen3.5 models, surpassing Qwen3.5-4B from $64$K onward and
approaching Qwen3.5-9B at $256$K. This trend indicates favorable
long-context scaling despite the relatively small activated parameter
budget of Turing-20B-A2B.

For Turing-20B-A2B, only the $256$K and $512$K results use
YaRN-based extrapolation from the $128$K-trained checkpoint. No
additional continued pretraining is performed for these evaluations.
The model maintains stable performance as the context is extended
beyond its native training range, with gradual degradation up to
$512$K. Together, these results suggest that progressive
long-context training provides robust scaling within the trained
context range, while YaRN effectively extends the usable context
window to substantially longer sequences without additional
parameter updates.

\subsection{Model Efficiency}
\begin{figure*}[t]
    \centering
    \includegraphics[width=1.0\textwidth]
    {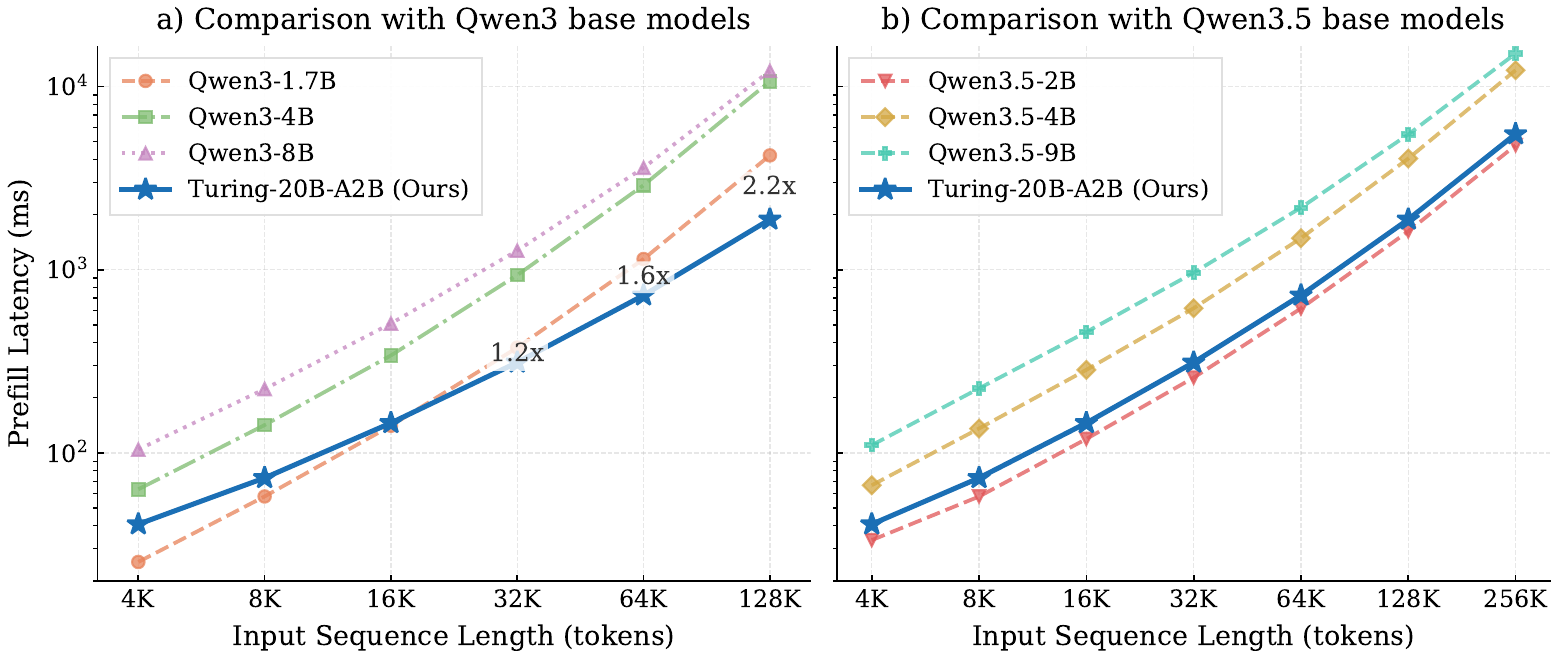}
    \caption{
        \textbf{Model-level prefill latency at different input
        sequence lengths compared with (a) Qwen3 and (b) Qwen3.5 base
        models.}
    }
    \label{fig:prefill_latency_scaling}
\end{figure*}

We further evaluate the prefill efficiency of Turing-20B-A2B and
representative Qwen base models across a wide range of input sequence
lengths. As shown in Figure~\ref{fig:prefill_latency_scaling}, we
compare Turing-20B-A2B with Qwen3 models from 4K to 128K tokens and
with Qwen3.5 models from 4K to 256K tokens. All latency experiments
are conducted on a single NVIDIA H800 GPU using FP16 precision and a
batch size of 1. The same compilation, input construction, warm-up,
timing, and synchronization protocol is applied to all evaluated
models. Complete benchmarking configurations are provided in
Appendix~\ref{app:efficiency_details}.

To focus on the sequence-length scaling of the model architecture, we
benchmark representative decoder layers and estimate model-level
prefill latency by aggregating the corresponding layer-level
measurements. For an input sequence of length $L$, the estimated
latency is computed as

\begin{equation}
    \widehat{T}_{\mathrm{prefill}}(L)
    =
    \sum_{k \in \mathcal{K}}
    N_k T_k(L),
    \label{eq:estimated_prefill_latency}
\end{equation}

where $\mathcal{K}$ denotes the set of distinct decoder-layer types,
$N_k$ is the number of layers of type $k$, and $T_k(L)$ is the
measured latency of a representative layer of that type at sequence
length $L$. For Turing-20B-A2B, Lightning Attention layers and
full-attention layers are measured separately and aggregated according
to their frequencies in the model.

As shown in Figure~\ref{fig:prefill_latency_scaling}, the latency
differences among the models are relatively modest at short sequence
lengths, whereas the advantage of Turing-20B-A2B becomes increasingly
pronounced as the context length grows. Compared with the Qwen3
baselines, Turing-20B-A2B becomes faster than the fastest baseline at
longer contexts, with the relative prefill speedup increasing from
approximately $1.2\times$ at 32K tokens to $1.6\times$ at 64K and
$2.2\times$ at 128K tokens.

A similar scaling advantage is observed when comparing against the
newer Qwen3.5 models. Turing-20B-A2B achieves substantially lower
prefill latency than Qwen3.5-4B and Qwen3.5-9B as the context length
increases, while remaining close to the substantially smaller
Qwen3.5-2B baseline. Notably, this favorable scaling persists up to
256K tokens, where the latency gap relative to the larger Qwen3.5
models becomes particularly pronounced.

The favorable long-context scaling of Turing-20B-A2B is enabled by
its attention-layer composition, in which Lightning Attention is used
in the majority of decoder layers while only a small number of layers
retain full attention. Consequently, the contribution of
quadratic-complexity attention is substantially reduced as the
sequence length grows, while periodic full-attention layers preserve
global token interactions. Together with the general capability
results, these measurements demonstrate a favorable trade-off among
model capability, activated computation, and long-context inference
efficiency.

\section{Ablation Studies}

\subsection{MoE Routing Strategy}
\label{sec:ablation_moe_routing}

We investigate the effect of the routing mechanism adopted in
Turing-20B-A2B through a controlled pretraining experiment. Specifically,
we compare Quantile Routing~\cite{su2025adaptive_moe} with Loss-Free
Routing~\cite{wang2024lossfree} using a 10B-parameter MoE model that
activates approximately 1.6B parameters per token. The model contains
128 routed experts and one shared expert. The Loss-Free Routing baseline
selects a fixed number of eight routed experts for each token, whereas
Quantile Routing dynamically varies the number of activated experts
across tokens while maintaining an average routed-expert budget of
eight. The shared expert is always activated and is not included in the
routed-expert budget.

Both variants are trained from scratch for approximately 100B tokens
using the same model architecture, training data, initialization, and
optimization configuration. The only difference between the two
variants is the routing mechanism. The update threshold for the
routing-control variables is set to 0.001 for both methods.

\begin{figure}[!t]
    \centering
    \includegraphics[width=0.8\textwidth]
    {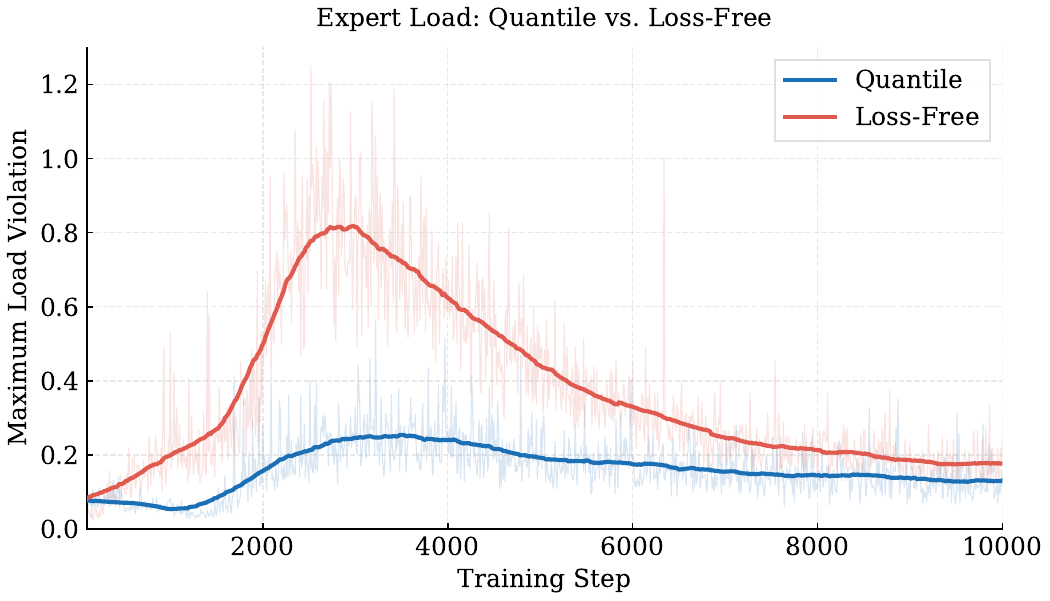}
    \caption{
        \textbf{Expert load balancing during pretraining.} We report the maximum
        load violation (\textit{MaxVio}) of the first MoE layer for
        Quantile Routing and Loss-Free Routing. The lightly shaded curves
        denote raw measurements, while the darker curves show smoothed
        trends. Lower values indicate better load balancing.
    }
    \label{fig:max_vio}
\end{figure}

\begin{table}[!t]
    \centering
    \caption{
        Early-stage pretraining performance of Quantile Routing and
        Loss-Free Routing. Both variants use the same model architecture,
        training configuration, and average routed-expert budget. The
        better result on each benchmark is shown in \textbf{bold}.
    }
    \label{tab:routing_ablation}
    \small
    \setlength{\tabcolsep}{4.5pt}
    \renewcommand{\arraystretch}{1.08}
    \begin{tabular*}{\linewidth}{
        @{\extracolsep{\fill}}
        lcccccccc
        @{}
    }
        \toprule
        \textbf{Routing}
        & \textbf{MMLU}
        & \textbf{ARC-C}
        & \textbf{TriviaQA}
        & \textbf{MATH}
        & \textbf{GSM8K}
        & \textbf{MBPP}
        & \textbf{BBH}
        & \textbf{HellaSwag} \\
        \midrule

        Loss-Free
        & 23.27
        & 23.73
        & 30.32
        & 19.72
        & \textbf{21.68}
        & 14.81
        & 27.39
        & 22.81 \\

        Quantile
        & \textbf{24.06}
        & \textbf{27.12}
        & \textbf{31.92}
        & \textbf{20.14}
        & 20.02
        & \textbf{18.78}
        & \textbf{28.18}
        & \textbf{25.18} \\

        \bottomrule
    \end{tabular*}
\end{table}

\paragraph{Expert load balancing.}
We first examine whether dynamic expert allocation adversely affects
expert load balancing. Following Loss-Free Routing
~\cite{wang2024lossfree}, we use the maximum load violation
(\textit{MaxVio}) to measure the largest deviation of an expert workload
from the target balanced workload. A lower MaxVio indicates better
worst-case load balancing.

Figure~\ref{fig:max_vio} reports the MaxVio of the first MoE
layer throughout training. The lightly shaded curves show the raw
measurements, while the darker curves represent their smoothed trends.
Quantile Routing exhibits substantially more stable expert utilization
during the early stage of training and maintains a lower MaxVio over
nearly the entire training trajectory. In contrast, the Loss-Free
Routing baseline experiences a pronounced increase in load violation
during the first several thousand steps and recovers only gradually as
training proceeds. These results indicate that, under the model and
training configuration considered here, Quantile Routing achieves more
stable expert load balancing despite allowing the number of activated
experts to vary across tokens.

\paragraph{Early-stage pretraining performance.}
We next evaluate the two routing variants at the same early-training
checkpoint on representative knowledge, reasoning, mathematics, and
coding benchmarks. All results use the same evaluation pipeline and
benchmark-specific configurations as the general capability evaluation
in Section~\ref{sec:overall_performance}, with full details provided in
Appendix~\ref{app:evaluation_details}.

As shown in Table~\ref{tab:routing_ablation}, Quantile Routing
outperforms Loss-Free Routing on seven of the eight evaluated benchmarks.
The improvements are observed across different capability categories,
with particularly clear gains on ARC-C, MBPP, and HellaSwag. GSM8K is
the only benchmark on which the Loss-Free Routing variant obtains a
higher score.

Because both variants use the same model architecture, training data,
optimization configuration, and average routed-expert budget, the
performance difference cannot be attributed to a larger average
computational cost. Instead, the results suggest that dynamically
allocating the available expert computation across tokens enables the
model to use the same average expert budget more effectively.

\begin{figure}[H]
    \centering
    \includegraphics[width=1.0\textwidth]
    {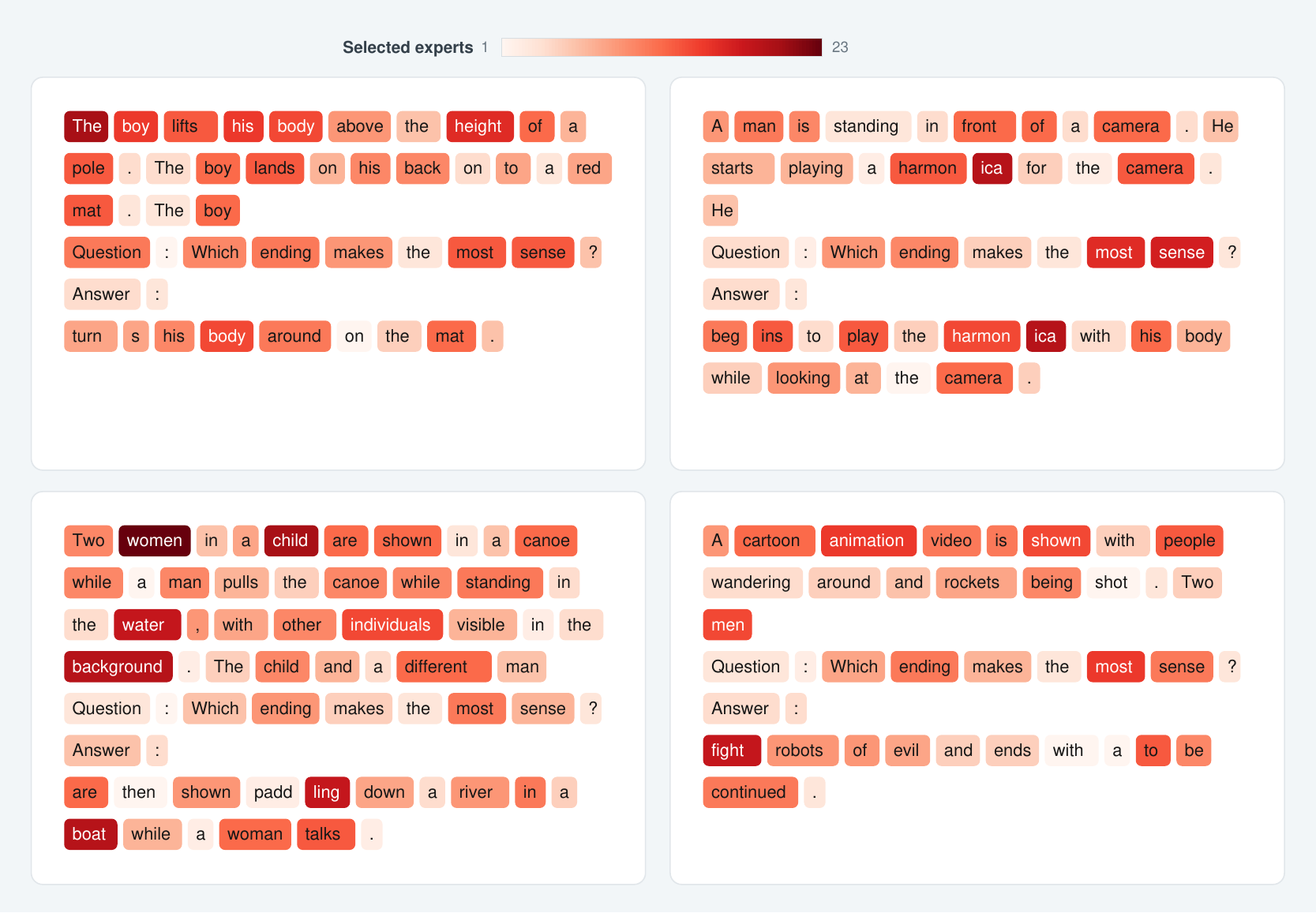}
    \caption{
        \textbf{Token-wise expert allocation produced by Quantile Routing on
        representative HellaSwag examples.} Darker tokens activate more
        routed experts. The expert count varies according to the
        contextual representation of each token, while maintaining an
        average routed-expert budget of eight. The always-active shared
        expert is excluded from the displayed counts.
    }
    \label{fig:token_expert_allocation}
\end{figure}

\paragraph{Token-wise expert allocation.}
A defining property of Quantile Routing is that different tokens may
activate different numbers of routed experts. To examine the learned
allocation pattern, we record the routed-expert count of each token on
examples from HellaSwag~\cite{zellers2019hellaswag}.
Figure~\ref{fig:token_expert_allocation} presents representative
visualizations, where darker tokens correspond to a larger number of
activated routed experts. The shared expert, which is always active,
is excluded from the displayed counts.

The qualitative examples in Figure~\ref{fig:token_expert_allocation}
show a clear context-dependent pattern in expert allocation.
Content-bearing and contextually informative tokens tend to activate
more routed experts, whereas punctuation marks and frequent function
words generally activate fewer. Different occurrences of the same
surface token may also receive different numbers of experts, indicating
that routing depends on contextual representations rather than token
identity alone. These observations are consistent with adaptive
computation, where more demanding tokens receive greater expert
capacity while relatively predictable tokens consume less computation.
Although expert count is not a direct measure of semantic importance,
the results suggest that dynamic top-$k$ Quantile Routing learns a
structured allocation of computation across tokens.

Overall, the controlled ablation shows that Quantile Routing provides
more stable expert load balancing and better early-stage pretraining
performance than Loss-Free Routing under the same average routed-expert
budget. The token-level analysis further illustrates that it allocates
computation in a context-dependent manner. These results motivate our
adoption of Quantile Routing in Turing-20B-A2B.

\subsection{Expert Capacity for Efficient Prefill}
\label{sec:capacity_ablation}

We evaluate expert-capacity control directly on Turing-20B-A2B from
two complementary perspectives: model capability and deployment
efficiency. For the capability study, we conduct reinforcement learning
using GSM8K as the sole training task and compare capacity-constrained
and dropless variants under the same Quantile Routing configuration. For the
efficiency study, we benchmark the MoE module of Turing-20B-A2B under
the corresponding deployment configurations.

\begin{table}[H]
    \centering
    \small
    \renewcommand{\arraystretch}{1.08}
    \setlength{\tabcolsep}{6pt}
    \caption{
        Effect of expert-capacity control during GSM8K-oriented
        reinforcement learning on Turing-20B-A2B. Both variants use
        Quantile Routing and are evaluated at RL step 399, differing
        only in whether CF $=1.25$ is applied during prompt prefill.
        The better result for each benchmark is highlighted in
        \textbf{bold}.
    }
    \label{tab:capacity_rl_ablation}

    \begin{tabular*}{0.85\textwidth}{
        @{\extracolsep{\fill}}
        lcccccc
        @{}
    }
        \toprule
        \textbf{Setting}
        & \textbf{MMLU}
        & \textbf{CMMLU}
        & \textbf{MMLU-Pro}
        & \textbf{BBH}
        & \textbf{GPQA}
        & \textbf{GSM8K} \\
        \midrule

        Dropless
        & \textbf{79.03}
        & \textbf{75.38}
        & \textbf{61.44}
        & 65.13
        & 31.31
        & \textbf{90.22} \\

        CF $=1.25$
        & 76.34
        & 70.48
        & 57.60
        & \textbf{66.02}
        & \textbf{34.34}
        & 88.93 \\

        \bottomrule
    \end{tabular*}
\end{table}

\begin{figure}[H]
    \centering
    \includegraphics[width=0.8\linewidth]
    {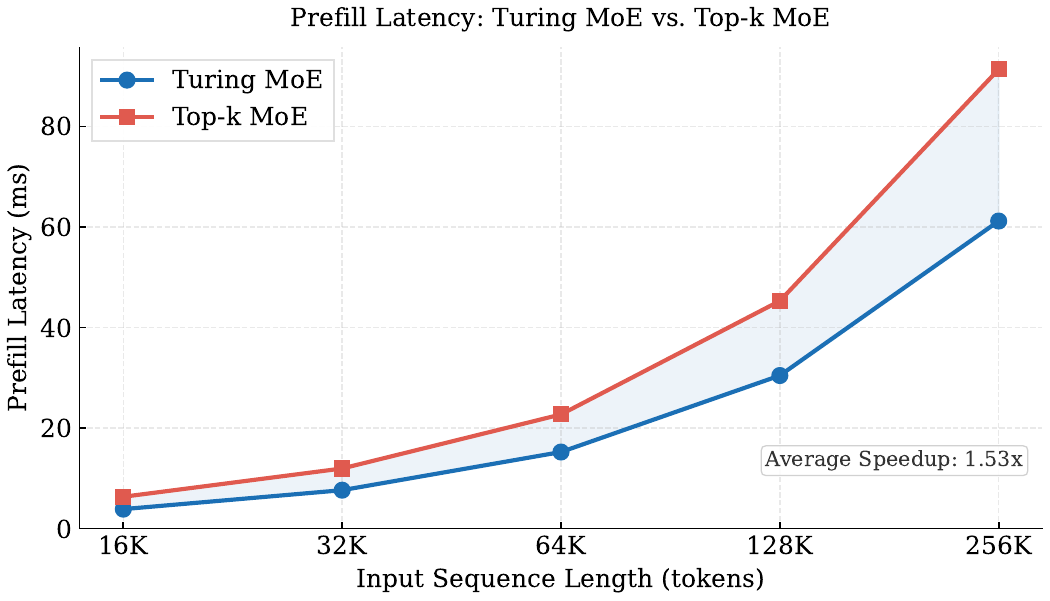}
    \caption{
        \textbf{MoE-module prefill latency comparison between Turing MoE and
        conventional top-$k$ MoE.} Turing MoE uses Quantile Routing with
        an expert capacity factor of $1.25$, while the top-$k$ baseline
        uses conventional dropless top-$k$ routing. Turing MoE achieves
        an average $1.53\times$ speedup over the evaluated context
        lengths.
    }
    \label{fig:moe_capacity_latency}
\end{figure}

\paragraph{Impact on model capability.}
We compare two reinforcement-learning runs of Turing-20B-A2B under
the same Quantile Routing configuration, differing only in whether the
expert-capacity constraint is enabled during prompt prefill. Both runs
are optimized on GSM8K and evaluated at the same RL step.

As shown in Table~\ref{tab:capacity_rl_ablation}, the
capacity-constrained variant remains broadly competitive with the
dropless setting, although some benchmark-level regressions are
observed. The dropless variant achieves higher scores on MMLU, CMMLU,
MMLU-Pro, and the directly optimized GSM8K task, whereas CF $=1.25$
performs better on BBH and GPQA. Importantly, the capacity constraint
does not lead to uniform capability degradation: the model retains
strong performance across the evaluated benchmarks while improving on
several reasoning tasks. These results indicate that capacity control
introduces a measurable but limited capability trade-off in exchange
for a more regular and efficient deployment-time expert-execution path.

\paragraph{Impact on prefill efficiency.}
We next evaluate the deployment-time efficiency of the MoE module in
Turing-20B-A2B. Turing MoE combines Quantile Routing with an expert
capacity factor of $1.25$, while the baseline uses conventional
dropless top-$k$ routing. The two implementations use the same model
dimensions and expert configuration and are evaluated under the same
benchmarking protocol. The top-$k$ baseline follows the fused-expert
execution paradigm commonly adopted by modern LLM inference engines,
where token--expert assignments are grouped by expert before fused
expert computation and weighted output aggregation.

Figure~\ref{fig:moe_capacity_latency} shows that Turing MoE consistently
reduces average MoE-module prefill latency across context lengths from
$16$K to $256$K. The advantage becomes increasingly pronounced as the
sequence length grows, with particularly large gaps at $128$K and
$256$K. Averaged over the evaluated sequence lengths, Turing MoE
achieves a $1.53\times$ speedup over the conventional dropless
top-$k$ MoE baseline.

The observed efficiency gain is consistent with the deployment
motivation for capacity-constrained routing. By combining Quantile
Routing with an explicit upper bound on per-expert workload, Turing MoE
enables a more regular expert-execution path and improves long-context
prefill efficiency. Together, the reinforcement-learning ablation and
the MoE-module latency measurements at the Turing-20B-A2B scale provide
complementary evidence that capacity-constrained prefill can
substantially improve deployment efficiency while retaining broadly
competitive downstream capability. Detailed benchmarking
configurations are provided in Appendix~\ref{app:moe_latency}.
\section{Applications and Practical Impact}
\label{sec:applications}

Turing-20B-A2B explores a \textbf{practical MoE scaling paradigm for
physical AI foundation models.} Physical AI systems require strong
general capabilities, long-context modeling, and efficient, stable
inference under strict deployment constraints. By combining dynamic
top-$k$ Quantile Routing with fine-grained experts, Turing-20B-A2B
allows different tokens to receive different amounts of expert
computation while maintaining balanced expert utilization and a
controlled average compute budget through a unified quantile-tracking
mechanism. This provides a practical path for scaling model capacity
and improving knowledge, reasoning, and long-context capabilities
without forcing every token to follow the same fixed expert-allocation
pattern. During prompt prefill, expert-capacity control further bounds
per-expert workloads, reducing workload fluctuations across experts and
making the execution pattern more regular and predictable. This is
particularly beneficial for deployment latency, where limiting
worst-case expert workload helps avoid large variations in execution
cost as routing distributions change. Controlled ablations show that
the capacity-constrained variant retains broadly competitive downstream
capability relative to the dropless setting while substantially
improving MoE prefill efficiency.

Hardware--model co-design is another central consideration behind
Turing-20B-A2B. The architecture favors simple, regular, and widely
supported computation patterns that are beneficial across a range of
edge and latency-sensitive devices, with the Turing chip serving as an
important target platform during model design. This consideration
motivates choices such as the Lightning-Attention-based hybrid
attention backbone, Dynamic Tanh in place of conventional
normalization, and fine-grained sparse expert computation. These
hardware-friendly designs help translate model-level efficiency into
practical improvements in inference latency and deployment stability.

These properties make Turing-20B-A2B suitable as a language foundation
model for a range of physical AI applications, including \textbf{VLA 2.0 for
autonomous driving, intelligent-cockpit and integrated
driving-and-parking systems, and embodied platforms such as the XPeng
Iron robot.} Although these applications differ in their downstream
interfaces and modalities, they share the need to process extended
histories of observations, states, instructions, and actions while
meeting strict latency and computation budgets. Turing-20B-A2B is
designed to provide a common foundation for such workloads, combining
strong base-model capability with long-context scalability and
deployment-oriented efficiency.
\section{Conclusion and Future Work}
\label{sec:conclusion_future_work}

This technical report presents Turing-20B-A2B, a 20B-parameter
Mixture-of-Experts language model that activates approximately 2B
parameters per token. The model combines dynamic top-$k$ Quantile
Routing, fine-grained experts, and a hybrid attention backbone
dominated by Lightning Attention. Quantile Routing enables
token-adaptive computation while maintaining balanced expert
utilization and a controlled average compute budget, and
capacity-constrained prompt prefill further regularizes expert
workloads for efficient deployment. Despite its compact active-parameter
budget, Turing-20B-A2B achieves base-model capability exceeding
Qwen3-8B and approaching Qwen3.5-9B, while maintaining strong
long-context performance and favorable prefill-latency scaling.
Progressive continued pretraining extends the native context window
from 4K to 128K, with training-free YaRN extrapolation further enabling
effective inference up to 512K. Overall, the results demonstrate an
effective balance among model capability, long-context scalability,
and practical inference efficiency for latency-sensitive physical AI
systems.

Looking forward, Turing-20B-A2B has so far been primarily validated as
a base language model, and its post-training capabilities remain under
active exploration. Future work will focus on supervised fine-tuning
and reinforcement-learning-based post-training, together with broader
evaluation on downstream tasks such as multi-turn interaction,
long-document understanding, long-context code understanding,
software engineering, and agent-oriented workloads. We also plan to
extend the model toward multimodal foundation models by integrating it
with TuringViT~\cite{wu2026turingvit}, providing a foundation for
applications such as autonomous driving and embodied intelligence that
require joint reasoning over long visual, language, state, and action
histories under practical latency constraints.
\section*{Contributors}
We sincerely thank every member of the team for their dedication and
valuable contributions. This work reflects our ongoing efforts to advance
VLM/VLA-related applications, and we hope that Turing-20B-A2B will play an
increasingly important role in this direction.

\textbf{Advisors}: Hang Zhang, HongGou Yang, Xianming Liu

\textbf{Project Lead}: Qiman Wu

\textbf{Contributors}: 
Yuheng Zhang\textsuperscript{*}
, Yizhao Wang\textsuperscript{*}
, Da Zhu\textsuperscript{*}
, Hua Zhou
, Yue He
, Jiahui Hu
, Shaman Tang
, Hanlin Chen
, Yuhua Wei
, Anhua Liu
, Shuang Su
, Rui Xin
, MingYuan Wang
, MingHao Li
, HaoJie Yang
, Siqi Liu
, Jianlei Zheng
, WeiChao Huang

\textsuperscript{*}Core contribution. 
{
    \begingroup
    \sloppy
    \small
    \bibliographystyle{turing_unsrtnat}
    \bibliography{main}
    \endgroup
}
\clearpage
\appendix
\begin{center}
    {\Large\bfseries Supplementary Material}
\end{center}
\vspace{1em}

\section{Base Model Evaluation Details}
\label{app:evaluation_details}

This appendix provides the detailed configurations used for the
general capability evaluation in
Section~\ref{sec:overall_performance}.
All experiments use the generation mode of OpenCompass. Each model
generates a textual response, which is subsequently processed by the
task-specific answer extractor and evaluator. Unless otherwise
specified, the maximum generation length is 1,024 tokens.

For fair comparison, all evaluated models use the same dataset
version and evaluation split, the same few-shot examples and example
ordering, identical prompt templates and chain-of-thought instructions,
the same generation configuration, and identical answer extraction and
scoring procedures. In particular, task-specific answer extraction is
kept consistent for benchmarks such as HellaSwag, WinoGrande, GPQA,
and DROP.

\begin{table*}[htbp]
    \centering
    \caption{
        Detailed evaluation configurations for the general capability
        benchmarks. ``CoT'' denotes chain-of-thought demonstrations.
    }
    \label{tab:evaluation_details}

    \small
    \renewcommand{\arraystretch}{1.10}

    \begin{tabular*}{0.96\textwidth}{
        @{\extracolsep{\fill}}
        lllll
        @{}
    }
        \toprule
        Benchmark
        & Split
        & \# Shots
        & Metric
        & Max Output \\
        \midrule

        \multicolumn{5}{c}{\textit{Knowledge Tasks}} \\
        \midrule

        MMLU
        & Test
        & 5-shot
        & Acc.
        & 1024 \\

        MMLU-Redux
        & MMLU-Redux 2.0
        & 5-shot
        & Acc.
        & 1024 \\

        CMMLU
        & Test
        & 5-shot
        & Acc.
        & 1024 \\

        C-Eval
        & Validation
        & 5-shot
        & Acc.
        & 1024 \\

        \midrule
        \multicolumn{5}{c}{\textit{Reasoning Tasks}} \\
        \midrule

        MMLU-Pro
        & Test
        & 5-shot CoT
        & Acc.
        & 1024 \\

        BBH
        & All 27 tasks
        & 3-shot CoT
        & EM
        & 4096 \\

        DROP
        & Validation subset
        & 0-shot
        & EM
        & 1024 \\

        WinoGrande
        & Dev
        & 5-shot
        & Acc.
        & 1024 \\

        HellaSwag
        & Validation
        & 0-shot
        & Acc.
        & 1024 \\

        \midrule
        \multicolumn{5}{c}{\textit{Math \& STEM Tasks}} \\
        \midrule

        ARC-C
        & Dev
        & 0-shot
        & Acc.
        & 1024 \\

        GPQA
        & Main
        & 5-shot CoT
        & Acc.
        & 1024 \\

        GSM8K
        & Test
        & 4-shot CoT
        & EM
        & 512 \\

        MATH
        & Test
        & 4-shot
        & EM
        & 1024 \\

        \bottomrule
    \end{tabular*}
\end{table*}

\paragraph{Knowledge benchmarks.}
MMLU is evaluated on the test set across 57 subjects, containing
14,042 questions in total. Five fixed examples, corresponding to
indices 0--4 of the development split of each subject, are used as
in-context demonstrations. The model generates an answer to an
A--D multiple-choice question, and the first valid option is extracted
before computing accuracy.

MMLU-Redux is evaluated using MMLU-Redux 2.0 over 57 subjects and
5,330 questions. It follows the same 5-shot prompting format as MMLU,
with five fixed development examples for each subject and generated
A--D answers evaluated by accuracy.

CMMLU is evaluated on its test split across 67 subjects and 11,582
questions. Five fixed examples from the development split are provided
for each subject. C-Eval is evaluated on its validation split across
52 subjects and 1,346 questions, also using five fixed development
examples. Both benchmarks use Chinese multiple-choice prompts and
accuracy after extracting the generated option.

\paragraph{Reasoning benchmarks.}
MMLU-Pro is evaluated on the test split, consisting of 12,032
questions across 14 categories. Five examples from the validation
split are provided with their chain-of-thought reasoning and final
answers. The model is prompted to reason step by step, after which
the final option is extracted from the generated response.

BBH is evaluated over all 27 BIG-Bench Hard subtasks, comprising
6,511 examples. Three chain-of-thought demonstrations are directly
embedded in each task-specific prompt. The target response is
generated following the instruction ``Let's think step by step.''
Multiple-choice and free-form BBH tasks use their corresponding
task-specific normalized answer matching procedures. The maximum
generation length is increased to 4,096 tokens to accommodate
intermediate reasoning.

DROP is evaluated zero-shot on a 6,114-example subset of the
validation set. Given a passage and question, the model directly
generates a free-form answer. The generated answer is normalized
using the DROP-specific post-processing procedure, and we report
exact match on this subset.

WinoGrande is evaluated on its 1,267-example development set.
Five fixed demonstrations with indices 0, 2, 4, 6, and 8 are drawn
from the \texttt{train\_xs} split. The two candidate expressions are
mapped to options A and B, and accuracy is computed after extracting
the generated option.

HellaSwag is evaluated zero-shot on all 10,042 examples of the
validation set. Each example contains a context and four candidate
continuations. The model generates the selected option, which is
processed using the same HellaSwag-specific answer extractor for
all compared models before accuracy is computed.

\paragraph{Math and STEM benchmarks.}
ARC-C is evaluated zero-shot on the ARC Challenge development split,
containing 295 examples. Each question is presented with four
candidate answers, and accuracy is computed from the extracted
A--D option.

GPQA is evaluated using the GPQA Main set, containing 448 questions,
rather than the GPQA Diamond subset. Five complete chain-of-thought
examples are directly embedded in the prompt. Models may generate
intermediate reasoning and are instructed to provide the final answer
in the required option format before accuracy is computed.

GSM8K is evaluated on its 1,319-example test set using four
chain-of-thought demonstrations embedded directly in the prompt.
Models generate a complete solution, after which the final numerical
answer is extracted and evaluated using exact match. The maximum
generation length is 512 tokens.

MATH is evaluated on all 5,000 examples of the test set with four
worked mathematical demonstrations. The model is allowed to generate
the full derivation, and the final mathematical answer is extracted
and evaluated using exact match.

\section{Model Efficiency Evaluation Details}
\label{app:efficiency_details}

This appendix describes the implementation and measurement protocol
used for the prefill-latency experiments reported in
Figure~\ref{fig:prefill_latency_scaling}. The benchmark measures the
forward-pass latency of representative decoder layers and aggregates
the measurements according to the layer composition of each model.
No tensor, pipeline, data, or expert parallelism is used during the
benchmark.

\paragraph{Hardware and software environment.}
All measurements are conducted on a single NVIDIA H800 GPU with
80\,GB-class memory. The benchmark environment uses CUDA 12.9,
PyTorch \texttt{2.8.0a0+5228986c39.nv25.06}, Triton 3.3.0,
Transformers 5.2.0, Flash Linear Attention 0.4.2,
FlashAttention \texttt{2.7.4.post1}, and
\texttt{causal-conv1d} \texttt{1.6.2.post1}. The models and inputs
use FP16 precision throughout the benchmark.

\paragraph{Input construction and runtime measurement.}
For a model with hidden size $H$ and sequence length $L$, the input
hidden states are sampled directly on the GPU from a standard normal
distribution with shape $[1,L,H]$. Position indices are constructed as
$[0,\ldots,L-1]$. All measurements use a batch size of 1 and run in
evaluation mode under \texttt{torch.no\_grad()}.

We compile each representative decoder layer with
\texttt{torch.compile} in \texttt{mode="reduce-overhead"}. Before
timing, we run a 32K-token forward pass to trigger the initial
compilation, followed by one untimed forward pass at each evaluated
sequence length to trigger any shape-specific compilation or
specialization. The reported latencies therefore exclude both initial
compilation and shape-specialization overheads.

Runtime is measured with \texttt{triton.testing.do\_bench}, using a
100\,ms warm-up period and a 5,000\,ms measurement period for each
sequence length. The reported value is the mean latency returned by
\texttt{do\_bench}. CUDA-event timing and synchronization are used
during measurement. The L2 cache is cleared before every timed
repetition by \texttt{do\_bench}, and the CUDA allocator cache is
cleared between sequence lengths.

\paragraph{Attention implementations.}
For the Hugging Face Qwen baselines, full-attention layers are
configured with \texttt{attn\_implementation="sdpa"}. On the H800 GPU,
PyTorch SDPA dispatches to the fused cuDNN native scaled-dot-product
attention kernel.

Qwen3.5 uses a hybrid architecture with model-defined linear-attention
layers and full-attention layers. In our benchmark, the linear-attention
layers retain the optimized execution path provided by the original
implementation. In particular, the short causal convolution in the
linear-attention block is executed through the optimized
\texttt{causal-conv1d} kernel rather than the generic PyTorch fallback,
while the linear-attention computation itself uses the corresponding
optimized kernel path. This avoids replacing the model's intended
high-performance implementation with unfused operators during latency
measurement. The Qwen3.5 full-attention layers use the same PyTorch
SDPA configuration as the other Qwen baselines.

The Lightning Attention layers in Turing-20B-A2B use a custom Triton
implementation adapted from Flash Linear Attention, with a block size
of 512 and decay enabled. Q, K, V, layer parameters, and outputs are
represented in FP16, while decay exponents are evaluated in FP32
inside the Triton kernels for numerical stability.

Turing-20B-A2B contains four causal full-attention layers implemented
with FlashAttention-2. The prefill path directly uses the known query
length for rotary-position construction, avoiding per-forward
GPU-to-CPU scalar synchronization or TorchDynamo graph breaks. The
KV-cache decoding path is not included in this benchmark.

Thus, all models are benchmarked using their respective optimized
attention implementations rather than reference or fallback paths.

\paragraph{Layer-level aggregation.}
Turing-20B-A2B contains 24 decoder layers, consisting of 20 Lightning
Attention layers and four full-attention layers. Five representative
decoder-layer types are benchmarked:

\begin{table}[htbp]
    \centering
    \caption{
        Representative decoder-layer types used to estimate the
        model-level prefill latency of Turing-20B-A2B.
    }
    \label{tab:turing_latency_aggregation}

    \small
    \renewcommand{\arraystretch}{1.08}

    \begin{tabular}{lc}
        \toprule
        \textbf{Layer Type} & \textbf{Multiplicity} \\
        \midrule
        Dense FFN + Lightning Attention
        & 1 \\
        Routing MoE + Lightning Attention
        & 4 \\
        Non-routing MoE + Lightning Attention
        & 15 \\
        Routing MoE + Full Attention
        & 2 \\
        Non-routing MoE + Full Attention
        & 2 \\
        \bottomrule
    \end{tabular}
\end{table}

Accordingly, the model-level prefill latency is estimated as

\begin{equation}
\begin{aligned}
\widehat{T}_{\mathrm{prefill}}(L)
={}&
T_{\mathrm{dense}}(L)
+ 4T_{\mathrm{routing,linear}}(L) \\
&+ 15T_{\mathrm{nonrouting,linear}}(L)
+ 2T_{\mathrm{routing,full}}(L) \\
&+ 2T_{\mathrm{nonrouting,full}}(L).
\end{aligned}
\end{equation}

For Qwen3-1.7B, the latency of one representative decoder layer is
multiplied by 28. For Qwen3-4B and Qwen3-8B, the representative
decoder-layer latency is multiplied by 36. Qwen3.5-4B and
Qwen3.5-9B each contain 24 linear-attention layers and eight
full-attention layers; the two layer types are measured independently
and aggregated according to these multiplicities.

\section{MoE Prefill Latency Benchmark Details}
\label{app:moe_latency}

The latency experiment in
Figure~\ref{fig:moe_capacity_latency} measures the forward latency of
a complete MoE module in isolation, excluding attention and
normalization overhead. We benchmark the MoE module at layer
index~$1$, corresponding to the first full MoE layer in the model.
The measurement therefore captures the complete inference path of each
MoE implementation, including routing, token selection or dispatch,
routed-expert computation, shared-expert computation, weighting, and
output aggregation.

We use the same hardware and software environment as the
model-efficiency experiments described in
Appendix~\ref{app:efficiency_details}. All measurements are performed on
a single NVIDIA H800 GPU with batch size $1$ and FP16 precision. For
each sequence length $L$, input hidden states are randomly generated
directly on the GPU with shape $[1,L,2048]$.
Figure~\ref{fig:moe_capacity_latency} reports results for sequence
lengths from $16$K to $256$K.

Both MoE variants share the same model dimensions and expert
configuration. The hidden dimension is $2{,}048$, with $256$ routed
experts and a nominal routing budget of $8$ experts per token. Each
routed expert has an intermediate dimension of $512$. Both variants
also contain the same shared expert with an intermediate dimension of
$2{,}048$.

Both MoE modules are evaluated through their dedicated inference
implementations and compiled with \texttt{torch.compile} using
\texttt{reduce-overhead} mode. Forward latency is measured using
\texttt{triton.testing.do\_bench}, with $100$\,ms of warm-up and a
$5{,}000$\,ms measurement window. The reported values correspond to the
mean forward latency returned by the benchmark. Model initialization,
input generation, attention, layer normalization, and host-to-device
input transfer are excluded from the timed region.

\paragraph{Turing MoE.}
Turing MoE uses Quantile Routing with sigmoid router scores,
expert-specific routing thresholds, and an expert capacity factor of
$\gamma=1.25$. For an input containing $B L$ tokens, the maximum
number of assignments processed by each routed expert is
\begin{equation}
    C
    =
    \left\lfloor
        \gamma \frac{BLk}{E}
    \right\rfloor,
    \label{eq:moe_latency_capacity}
\end{equation}
where $B=1$, $k=8$, and $E=256$ in our experiments. For each expert,
the router scores over the input tokens are ranked and at most the
highest-scoring $C$ assignments are retained. The resulting bounded
per-expert token blocks are gathered into regular expert-wise tensors
and processed using batched matrix multiplications for the two expert
MLP projections, followed by weighted aggregation back to the original
token positions.

\paragraph{Top-$k$ MoE baseline.}
The baseline uses conventional top-$k$ routing with dropless expert
execution. All $k$ selected token--expert assignments are retained, so
the number of tokens processed by an individual expert varies with the
input routing distribution.

To provide a representative deployment baseline, we implement the
dropless top-$k$ MoE following the fused-expert execution paradigm
commonly adopted by modern LLM inference engines such as vLLM and
SGLang~\cite{kwon2023vllm,zheng2024sglang}. Rather than executing
experts through a naive per-expert loop, token--expert assignments are
first grouped according to their selected experts. The resulting
expert-wise token groups are then processed using fused kernels for the
two expert MLP projections, gated activation, routing-weight scaling,
and output aggregation.

Concretely, the implementation sorts token--expert assignments by
expert and computes the corresponding expert offsets. The fused MoE
kernels then execute the first projection together with the gated
activation, followed by the second projection, routing-weight scaling,
and scatter-add to the original token positions. This execution path
supports variable per-expert workloads while avoiding the overhead of
a naive expert-by-expert implementation.

Both implementations use the same model dimensions, expert
configurations, input shapes, numerical precision, compilation settings,
and timing protocol, differing only in routing and expert execution. We
compare the complete Turing MoE design---Quantile Routing with
capacity-constrained execution---against an optimized dropless top-$k$
baseline using the fused-expert paradigm of modern inference systems.
The benchmark neither isolates the capacity factor nor includes
expert-parallel all-to-all communication; it measures single-GPU local
MoE execution rather than end-to-end distributed latency.

\end{document}